\documentclass[10pt,journal,compsoc]{IEEEtran}
\usepackage{amsmath,amsfonts}

\ifCLASSOPTIONcompsoc
\else
\fi

\usepackage{lipsum}
\usepackage{algorithmic}
\usepackage{algorithm}
\usepackage{array}
\usepackage[caption=false,font=normalsize,labelfont=sf,textfont=sf]{subfig}
\usepackage{textcomp}
\usepackage{stfloats}
\usepackage{url}
\usepackage{verbatim}
\usepackage{graphicx}
\usepackage{cite}
\usepackage{mathrsfs}
\usepackage{multirow}
\usepackage{booktabs}
\usepackage{ragged2e}
\usepackage{setspace}
\usepackage[pagebackref=true,breaklinks=true,citecolor=blue,linkcolor=blue,urlcolor=blue,colorlinks,bookmarks=false]{hyperref}
\usepackage[table]{xcolor}
\usepackage{amssymb} 
\usepackage{pifont}

\definecolor{bestcolor}{HTML}{EADCF8}
\definecolor{secondcolor}{HTML}{FFF2CC}

\newcommand{\best}[1]{\cellcolor{bestcolor}\textbf{#1}}
\newcommand{\second}[1]{\cellcolor{secondcolor}#1}
\newcommand{\cmark}{\textcolor{green!60!black}{\ding{52}}} \newcommand{\xmark}{\textcolor{red!70!black}{\ding{56}}} 
\makeatletter
\newcommand{\thickhline}{%
	\noalign {\ifnum 0=`}\fi \hrule height 1pt
	\futurelet \reserved@a \@xhline
}
\makeatother

\begin{document}

\title{More Than Where You Are: Learning Semantics, Structure, and Geometry from Cross-View Localization}

\author{Mao Chen, Xiangkai Zhang, Zhiyong Liu, \textit{Senior Member, IEEE}, \\Chuankai Liu, and Xu Yang, \textit{Senior Member, IEEE}
\thanks{Mao Chen, Xiangkai Zhang, Zhiyong Liu, and Xu Yang are with the State Key Laboratory of Multimodal Artiﬁcial Intelligence Systems, Institute of Automation, Chinese Academy of Sciences, and also with the School of Artiﬁcial Intelligence, University of Chinese Academy of Sciences. \protect\\
Chuankai Liu is with the Beijing Aerospace Control Center. \protect\\
(Corresponding author: Xu Yang.)}% <-this % stops a space
%\thanks{Manuscript received April 19, 2021; revised August 16, 2021.}
}

\IEEEtitleabstractindextext{%

\begin{abstract} 
\justifying
Consistent cross-view understanding under extreme viewpoint changes is essential for spatial intelligence, as it enables models to recognize the same scene across extreme viewpoint gaps.
Cross-view localization naturally provides a promising pathway toward this ability, as it requires a model to align ground-view imagery with geo-referenced satellite-view imagery despite drastic appearance changes to estimate camera poses.
Recent visual foundation models have made this long-standing localization problem increasingly feasible by providing rich 2D representations for cross-view matching.
However, we argue that cross-view localization should not be viewed merely as 2D matching or pose estimation. 
In this work, we revisit cross-view localization as more than pose estimation and investigate how it can help the model develop consistent cross-view understanding under extreme viewpoint changes, including stable semantics, reliable structure, and transferable geometry.
We identify three key limitations of existing methods that prevent them from achieving this. 
They usually lack explicit 3D grounding, rely on strict point-wise matching that can weaken semantic consistency, and learn from an absolute objective that provides limited guidance for geometric reasoning. 
To address these limitations, we propose CROSS, a unified cross-view localization framework built upon 3D-grounded alignment, structure-aware matching, and hypothesis ranking.
This formulation makes structure learning an intrinsic requirement of localization, encourages semantic representations to remain stable beyond rigid one-to-one correspondence, and enables the model to acquire transferable geometry that generalizes beyond seen viewpoint transformations.
Extensive experiments on the KITTI and VIGOR datasets show that CROSS achieves state-of-the-art performance in cross-view localization.
More importantly, CROSS effectively learns stable semantics, reliable structure, and transferable geometry across extremely different viewpoints.
The source code is available at \href{https://github.com/maochen-casia/CROSS}{https://github.com/maochen-casia/CROSS}.
\end{abstract}


\begin{center}
    \includegraphics[width=0.9\textwidth]{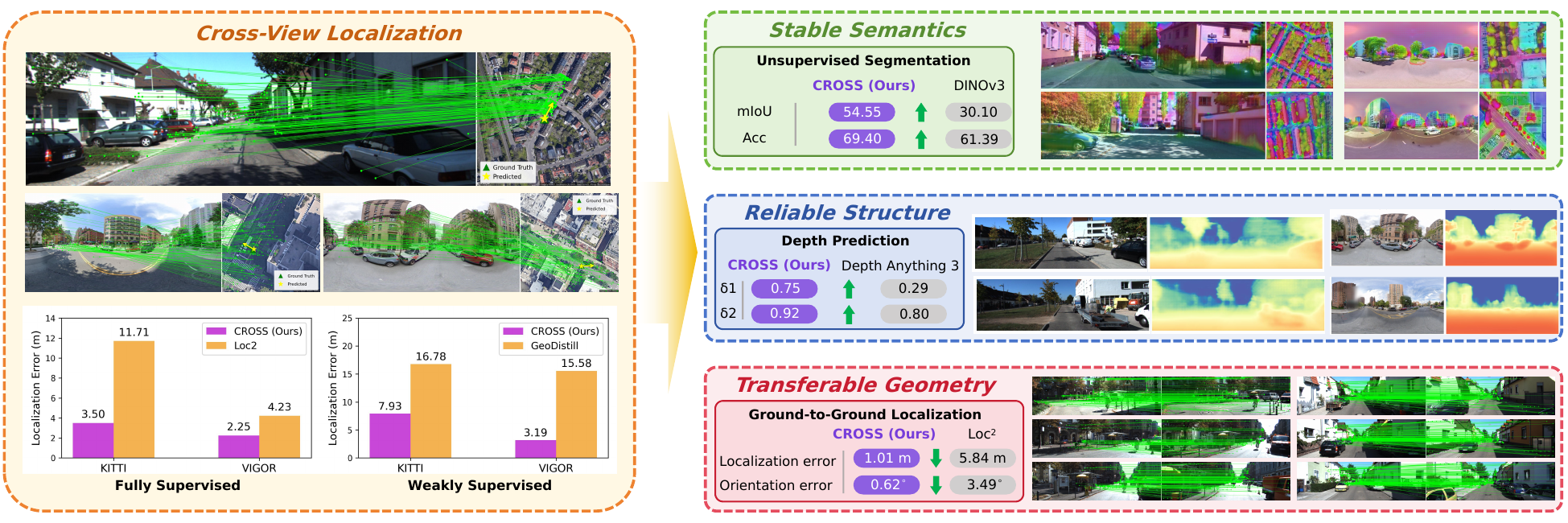}
    \setcounter{figure}{0}
    \refstepcounter{figure}
    \label{fig:teaser}
    \vspace{1mm}
    
    \begin{minipage}{0.92\textwidth}
    \footnotesize
    Fig.~\thefigure. CROSS revisits cross-view localization as more than pose estimation.
    Beyond achieving superior localization performance under both fully and weakly supervised settings, CROSS acquires stable semantics, reliable structure, and transferable geometry, outperforming strong foundation models and localization baselines across multiple dimensions.

    \end{minipage}
\end{center}
}

\maketitle

\IEEEdisplaynontitleabstractindextext
\IEEEpeerreviewmaketitle
\IEEEraisesectionheading{\section{Introduction} \label{sec:introduction}}

\IEEEPARstart{C}{onsistent} cross-view understanding is essential for spatial intelligence, as it enables agents to recognize the same scene from different viewpoints~\cite{chen2024spatialvlm,cheng2024spatialrgpt,wang2025cut3r,wang2025moge}. 
Recent foundation models have greatly advanced this capability under small viewpoint changes.
Semantic foundation models such as DINO~\cite{caron2021dino,oquab2024dinov2,simeoni2025dinov3} learn rich dense representations that capture object-level and region-level semantics, while 3D foundation models such as Depth Anything 3 (DA3)~\cite{lin2025da3} recover plausible scene structure and camera geometry.
However, these capabilities are still largely developed from observations within a single view or across relatively small viewpoint changes~\cite{wang2024dust3r,leroy2024mast3r,wang2025vggt,keetha2026mapanything}.
When the viewpoint gap becomes extreme, such as between ground-view imagery and overhead satellite-view imagery, current visual foundation models still struggle to achieve consistent cross-view understanding.
For example, semantic representations may be reliable within each individual view but fail to establish consistent correspondences across drastically different viewpoints, while structure and geometry that appear locally plausible may not remain consistent across such views.
Addressing this limitation is particularly important for spatial intelligence, as it allows models to reason despite drastic appearance variations and develop more generalizable scene understanding.

Humans naturally develop such an ability.
During self-localization, for example, when observing a scene from the ground, we can associate ground-level observations with an overhead map, forming a consistent understanding of the same scene across drastically different viewpoints.
Inspired by this, we investigate \textbf{whether cross-view localization can serve as a learning framework for acquiring consistent cross-view understanding under extreme viewpoint changes, including stable semantics, reliable structure, and transferable geometry}.

\begin{figure}
    \centering
    \includegraphics[width=0.95\columnwidth]{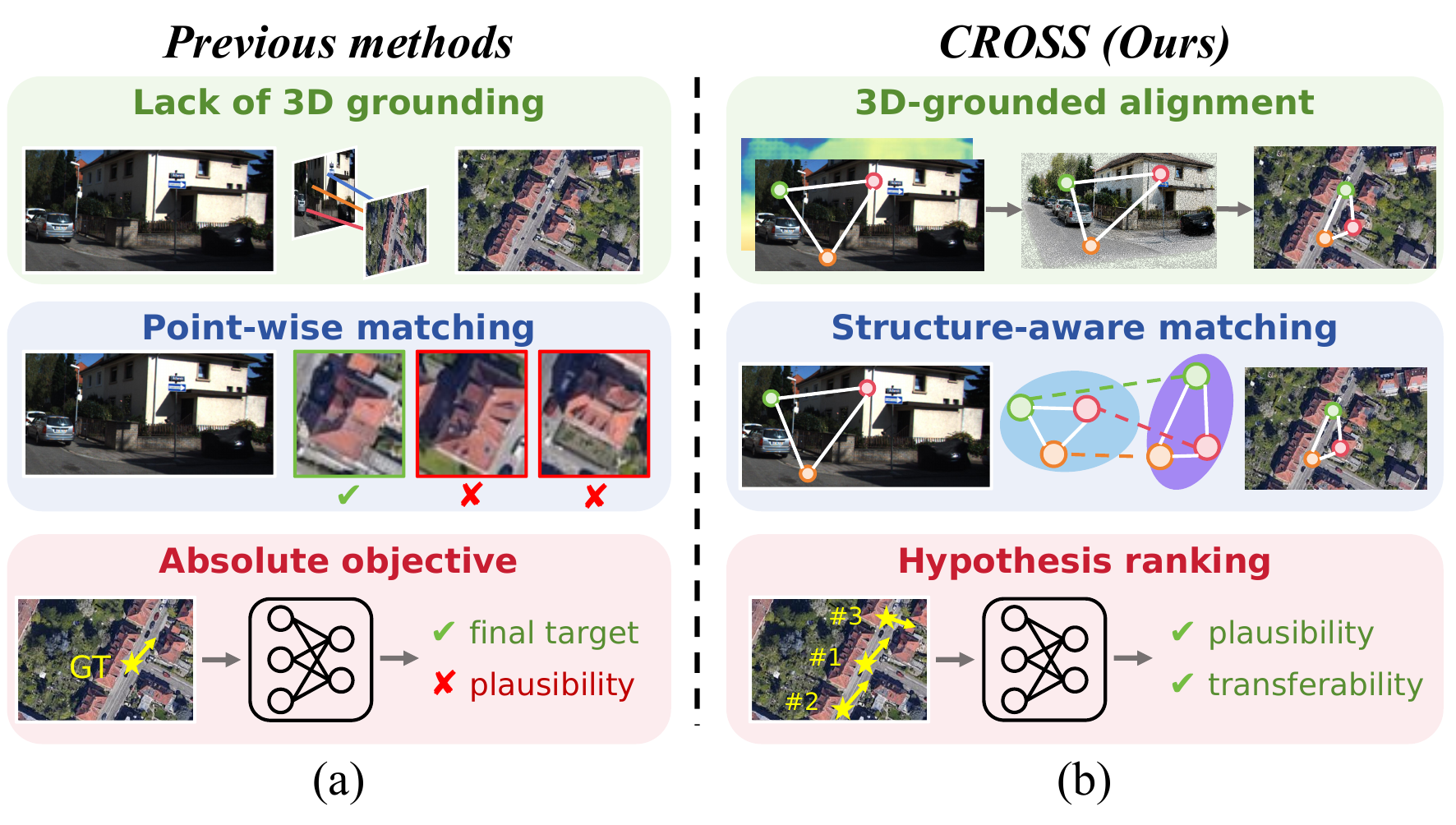}
    \caption{Comparison between previous methods and our proposed CROSS in learning semantics, structure, and geometry from cross-view localization. (a) Most existing methods establish 2D-2D correspondences through point-wise matching and learn from an absolute objective. (b) CROSS introduces 3D-grounded alignment and structure-aware matching to evaluate holistic cross-view consistency, and learns through hypothesis ranking under a unified framework.}

    \label{fig:intro}
\end{figure}

Cross-view localization naturally provides a promising pathway toward consistent cross-view understanding under extreme viewpoint changes, as it requires a model to align ground-view imagery with geo-referenced satellite-view imagery to estimate camera poses~\cite{lin2015wherecnn,workman2015widearea,vo2016localizing,liu2019lending,zhu2021vigor}.
With the rapid development of visual foundation models, recent studies have shown that establishing sparse correspondences across drastically different viewpoints is becoming increasingly feasible~\cite{xia2025fg2,xia2026loc2}.
By exploiting the rich 2D representations provided by foundation models, these methods typically formulate localization as a two-stage pipeline, where sparse 2D correspondences are first established between ground-view and satellite-view images, and the camera pose is then estimated from the predicted matches.
This sparse matching paradigm resembles human self-localization, where a few distinctive landmarks can often be sufficient for accurate localization.
However, the two-stage pipeline decouples correspondence from pose, overlooking the fact that cross-view localization is fundamentally a 3D reasoning problem, in which correspondences and the camera pose are intrinsically coupled~\cite{shi2026planerectrpp}.
To be more specific, a correspondence is meaningful only when it is supported by a valid pose hypothesis, while a pose can be reliably determined only when the induced correspondences are globally consistent.
Such decoupling limits the learning potential of cross-view localization for acquiring consistent cross-view understanding under extreme viewpoint changes. 
In particular, it poses three technical limitations for learning stable semantics, reliable structure, and transferable geometry, as shown in Fig.~\ref{fig:intro}(a).

First, existing formulations lack explicit 3D grounding, which limits their ability to support structure learning~\cite{lentsch2023slicematch,xia2024ccvpe,xia2025fg2,xia2026loc2}.
Most methods learn correspondences directly in the 2D image domain, without requiring the model to reconstruct, infer, or reason about the underlying 3D scene structure.
As a result, localization is often solved primarily through appearance matching, providing limited incentive for the model to learn reliable scene structure.

Second, most methods enforce strict point-wise matching across views~\cite{xia2025fg2,xia2026loc2}, encouraging the model to distinguish visually similar objects in an exclusive manner, which can hinder the learning of stable semantics.
Cross-view localization naturally exhibits many-to-many semantic consistency.
For example, a house in the ground-view image should not be semantically associated with only one house in the satellite-view image, but rather with multiple semantically consistent house instances.
Enforcing rigid one-to-one point-wise matching can bias the representation toward instance discrimination and weaken semantic consistency across viewpoints.

Third, existing learning objectives provide limited supervision for learning transferable geometry.
They often treat localization as absolute pose prediction, rather than evaluating which pose hypothesis best explains cross-view geometric consistency~\cite{lentsch2023slicematch,xia2024ccvpe,shi2023ggcvt,wang2023hcnet,song2023denseflow}.
In this formulation, ground-truth poses serve only as final targets, offering limited guidance on why a pose is plausible or how it differs from alternative hypotheses.
This limitation becomes more pronounced under weak supervision, where previous methods often decouple orientation and translation to exploit partially available pose labels~\cite{shi2024g2sweakly,xia2024adapting,tong2025geodistill}.
Such decoupling ignores the inherent geometric dependency between orientation and translation, preventing the model from evaluating the plausibility of coupled pose hypotheses as a whole.
Without explicitly reasoning about coupled pose hypotheses through cross-view consistency, existing methods struggle to learn coherent camera geometry and often fail to generalize beyond the trained ground-to-satellite setting.

To address these limitations, we propose CROSS, a unified framework for developing consistent cross-view understanding and learning stable semantics, reliable structure, and transferable geometry from cross-view localization.
Rather than treating localization as independent 2D matching followed by pose estimation, CROSS introduces 3D-grounded alignment and structure-aware matching to evaluate the cross-view consistency of pose hypotheses, and learns through hypothesis ranking, as shown in Fig.~\ref{fig:intro}(b).

First, CROSS performs 3D-grounded alignment to learn reliable structure.
Given a ground-view image, sparse visual representations are lifted into 3D using predicted metric-scale depth and then projected onto the satellite-view image under each pose hypothesis.
This establishes a 2D-3D-2D alignment pathway, where a hypothesis can be verified only when the predicted ground-view structure is consistent with the metric layout of the satellite view.
Therefore, structure is not learned merely to be locally plausible within a single viewpoint, but to be cross-view consistent as a representation of the same world.

Second, CROSS learns stable cross-view semantics through structure-aware matching.
Instead of enforcing brittle one-to-one correspondences between viewpoints, CROSS aggregates matching evidence over the holistic structure induced by each pose hypothesis.
This allows the model to evaluate whether a set of scene elements forms a globally plausible alignment, rather than requiring every local feature to match independently.
For example, a house in the ground view may be matched to a different but semantically similar house in the satellite view under an incorrect pose hypothesis.
Such a local ambiguity is not overly penalized, since the model can accumulate contradictory evidence from the holistic scene structure.
As a result, semantic representations are encouraged to remain stable across extreme viewpoints, rather than collapsing into instance-specific biases.

Third, CROSS learns transferable geometry by ranking pose hypotheses according to cross-view consistency.
Each pose hypothesis specifies the observer-dependent camera geometry and determines how the predicted structure is projected across viewpoints.
The model therefore learns not only what the target pose is, but also why it is plausible, since a valid pose should produce semantically, structurally, and geometrically consistent alignment across views.
Under full supervision, the ground-truth pose is ranked above perturbed hypotheses.
Under weak supervision, the missing pose dimensions are treated as latent variables within the same hypothesis space and optimized through their consistency with the known pose dimensions.
This unified formulation preserves the coupling between orientation and translation, thereby supporting both fully and weakly supervised training.
By ranking hypotheses through cross-view consistency, CROSS acquires transferable geometry that generalizes beyond the trained viewpoint transformation.

The contributions of this work are summarized as follows:
\begin{itemize}
    \item Beyond pose estimation, we revisit cross-view localization as a learning framework for acquiring consistent cross-view understanding under extreme viewpoint changes, including stable semantics, reliable structure, and transferable geometry. 
    \item We propose CROSS, a unified cross-view localization framework that introduces 3D-grounded alignment and structure-aware matching to support reliable structure learning and stable semantic representation.
    \item We introduce a pose hypothesis ranking objective for transferable geometry learning. By ranking pose hypotheses according to their induced cross-view consistency, CROSS naturally supports both fully and weakly supervised settings and learns transferable geometry beyond the trained transformation.
    \item Extensive experiments demonstrate that CROSS achieves state-of-the-art localization performance and effectively learns stable semantics, reliable structure, and transferable geometry from cross-view localization.
\end{itemize}

\section{Related Work}
\label{sec:related_work}

We discuss related methods in three aspects: cross-view localization, semantic learning, and structural and geometric learning.

\vspace{1mm}
\noindent \textbf{Cross-View Localization.}
Cross-view localization requires a model to align ground-view imagery with geo-referenced satellite-view imagery to estimate camera poses, thus providing a promising pathway toward consistent cross-view understanding under extreme viewpoint changes.
Early studies commonly formulated this problem as image retrieval, where a ground-view query is matched against a database of satellite-view images using global descriptors or learned cross-view embeddings~\cite{lin2015wherecnn,workman2015widearea,vo2016localizing,liu2019lending,shi2020optimal,yang2021l2ltr,zhu2021vigor,hu2018cvmnet,shi2019safa,zhu2022transgeo}.
Recent fine-grained localization methods move beyond global retrieval and aim to estimate 3-DoF camera poses, including translation and orientation, by establishing dense or sparse alignments between ground-view and satellite-view images~\cite{shi2022beyond,xia2024ccvpe,shi2023ggcvt,wang2023hcnet,song2023denseflow,xia2025fg2,xia2026loc2}.

\noindent \textit{Dense Matching Methods.}
Motivated by the strong perception capability of bird's-eye-view (BEV) representations in 3D scene understanding~\cite{huang2021bevdet,li2022bevformer,li2023bevdepth,yang2023bevformerv2}, early fine-grained cross-view localization methods commonly reduce the drastic viewpoint gap by transforming ground-view observations into dense BEV or map-aligned representations, followed by dense alignment with satellite imagery~\cite{shi2023ggcvt,wang2023hcnet,song2023denseflow,wang2024ovc,sarlin2023orienternet,sarlin2023snap}.
These methods provide a spatially organized intermediate representation between ground and aerial views, but they also inherit the limitations of dense BEV representations.
Perspective-to-BEV transformation may compress height information, blur fine-grained details, and introduce redundant features, making alignment computationally expensive and less interpretable.

\noindent \textit{Sparse Matching Methods.}
Recently, sparse matching methods have emerged as a more favorable alternative to dense BEV alignment.
Rather than correlating dense feature maps, these methods establish sparse correspondences between ground-view and satellite-view images and estimate the camera pose from the predicted matches.
FG$^2$ formulates fine-grained cross-view localization as feature matching between a ground-derived point plane and an aerial point plane, improving interpretability by tracing which local features contribute to pose estimation~\cite{xia2025fg2}.
Loc$^2$ further learns ground-to-satellite correspondences with monocular depth priors, enabling pose estimation from depth-lifted local correspondences~\cite{xia2026loc2}.
These methods mark an important shift from implicit dense matching toward explicit and interpretable sparse matching.

Nevertheless, most sparse matching methods still separate correspondence establishment from pose estimation, rely on strict point-wise matching, and learn from an absolute objective.
As a result, despite their improved localization effectiveness and efficiency, they do not fully exploit the potential of cross-view localization for acquiring consistent cross-view understanding under extreme viewpoint changes.
In contrast, CROSS introduces 3D-grounded alignment to learn reliable structure, structure-aware matching to develop stable semantics, and hypothesis ranking to acquire transferable geometry.

\vspace{1mm}
\noindent \textbf{Semantic Learning.}
Recent vision foundation models have demonstrated remarkable capabilities in semantic representation learning through large-scale pretraining.
Self-supervised methods such as DINO~\cite{caron2021dino} show that object- and part-level semantics can emerge from visual pretraining without dense annotations, while DINOv2~\cite{oquab2024dinov2} and DINOv3~\cite{simeoni2025dinov3} further scale this paradigm to obtain strong general-purpose features across diverse visual domains.
Vision-language models also learn transferable semantics from image-text supervision and enable broad zero-shot recognition~\cite{radford2021clip,jia2021align,li2021albef,zhai2022lit,li2022blip,li2023blip2,zhai2023siglip,tschannen2025siglip2}.
These foundation models provide powerful semantic priors and have been widely adopted as generic visual encoders for downstream tasks.

However, the semantics learned by these models are not explicitly constrained to remain consistent across extreme viewpoint changes. 
Most pretraining objectives operate on single images, image-text pairs, or visually related views, rather than paired observations of the same 3D scene from drastically different viewpoints.
As a result, their features may capture strong intra-view semantics while still exhibiting inconsistent cross-view correspondences between ground-view and satellite-view imagery.
In contrast, CROSS uses cross-view localization itself as supervision for semantic learning, encouraging semantic representations to remain stable across viewpoints and directly support cross-view reasoning.

\vspace{1mm}
\noindent \textbf{Structural and Geometric Learning.}
Recent vision foundation models have also made rapid progress in learning scene structure and camera geometry from large-scale data.
Monocular depth models, such as Depth Anything~\cite{yang2024depthanything}, Depth Anything V2~\cite{yang2024depthanythingv2}, and UniDepthV2~\cite{piccinelli2026unidepthv2}, learn strong depth priors that generalize across diverse image domains, providing a practical foundation for single-view structure understanding.
Recent models further move beyond independent depth prediction toward unified 3D perception.
DUSt3R~\cite{wang2024dust3r} formulates unconstrained stereo reconstruction as dense point-map regression, enabling depth, matching, and camera estimation without known calibration or poses.
Following this direction, MASt3R~\cite{leroy2024mast3r} strengthens image matching with 3D-grounded representations, and VGGT~\cite{wang2025vggt} unifies key geometric predictions, including camera parameters, depth maps, point maps, and point tracks, in a feed-forward framework.
MapAnything~\cite{keetha2026mapanything} further extends this trend toward universal metric 3D reconstruction by accepting flexible geometric inputs, while Depth Anything 3~\cite{lin2025da3} predicts spatially consistent geometry by introducing a unified depth-ray representation.
These works show that large-scale pretraining can substantially enhance unified spatial understanding.

Despite these advances, existing foundation models are mainly designed to recover structure and geometry within a single viewpoint or across relatively small viewpoint changes.
Their predictions can be locally plausible under standard perspective geometry, but are not explicitly optimized to explain consistency across extremely different viewpoints, such as ground-view and satellite-view imagery.
In contrast, CROSS learns reliable structure and transferable geometry from cross-view localization by evaluating cross-view consistency of hypotheses, thereby generalizing beyond the trained viewpoint transformation.

\section{Proposed Method}
\label{sec:method}

\begin{figure*}[!t]
    \centering
    \includegraphics[width=0.95\textwidth]{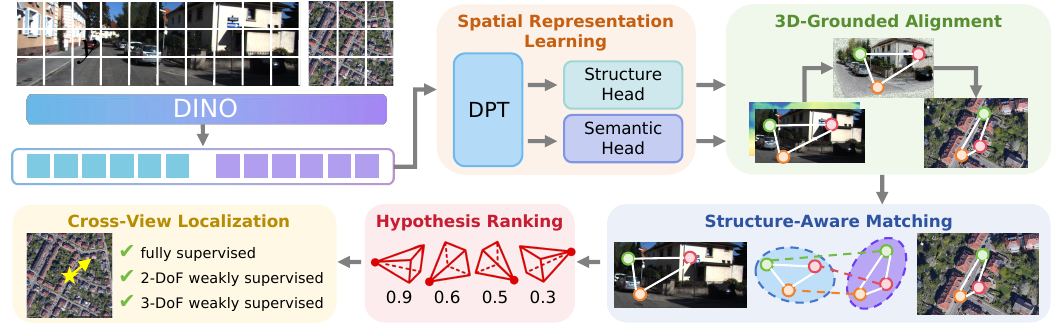}
    \caption{Overall framework of CROSS. The spatial representation learning module first uses a structure head and a semantic head to predict the structure and semantic embeddings. Then, sparse correspondences are established by 3D-grounded alignment, and cross-view consistency is evaluated through structure-aware matching. Finally, the model is trained with hypothesis ranking, which supports both fully and weakly supervised training within a unified framework.}
    \label{fig:framework}
\end{figure*}

Given a ground-view image $\mathbf{I}_g$ with an initial noisy 3-DoF camera pose $(\tilde{x}, \tilde{y}, \tilde{\theta})$, where $(\tilde{x}, \tilde{y})$ denotes the horizontal translation and $\tilde{\theta}$ denotes the orientation angle, together with a geo-referenced satellite-view image $\mathbf{I}_s$, CROSS estimates the refined 3-DoF pose $(\hat{x}, \hat{y}, \hat{\theta})$ of the ground-view camera.
The overall framework is illustrated in Fig.~\ref{fig:framework} and consists of four main components.
First, \textbf{spatial representation learning} predicts scene structure and semantic embeddings.
Second, \textbf{3D-grounded alignment} introduces a 2D-3D-2D pathway that encourages the model to learn reliable structure.
Third, \textbf{structure-aware matching} aggregates consistency evidence over the holistic structure to guide the learning of stable semantics.
Finally, \textbf{hypothesis ranking} enables CROSS to learn from both fully and weakly supervised training within a unified framework and develop transferable geometric understanding.

\subsection{Spatial Representation Learning}

The spatial representation learning stage aims to extract rich semantic and structural features from both ground-view and satellite-view images. 

A frozen DINO encoder~\cite{oquab2024dinov2,simeoni2025dinov3} $\Phi_{\mathrm{DINO}}$ first extracts intermediate visual tokens, which are then processed by a shared DPT~\cite{ranftl2021dpt} decoder $\Phi_{\mathrm{DPT}}$ to produce multi-scale feature maps at strides $\ell \in \{1,2,4,8\}$:
\begin{equation}
    \mathbf{F}_g^\ell = \Phi_{\mathrm{DPT}}(\Phi_{\mathrm{DINO}}(I_g)), \quad
    \mathbf{F}_s^\ell = \Phi_{\mathrm{DPT}}(\Phi_{\mathrm{DINO}}(I_s)).
\end{equation}

Unlike previous methods, we employ a shared decoder $\Phi_{\mathrm{DPT}}$ across viewpoints for both structural and semantic learning, encouraging the model to acquire a robust and generic spatial representation.

\subsubsection{Structural Learning}

To capture the structural layout of the scene, CROSS applies a \textit{structure head} to ground-view features and predicts metric-scale depth maps.

For each scale $\ell \in \{1,2,4,8\}$, the corresponding ground-view feature map $\mathbf{F}_g^\ell \in \mathbb{R}^{C \times H \times W}$ is processed by a Multi-Layer Perceptron (MLP) to produce depth distribution logits $\tilde{\mathbf{D}}_{g}^\ell \in \mathbb{R}^{D \times H \times W}$:
\begin{equation}
    \tilde{\mathbf{D}}_g^\ell = \Phi_{\mathrm{depth}}(\mathbf{F}_g^\ell),
\end{equation}
where the $D$ channels of $\tilde{\mathbf{D}}_{g}^\ell$ represent logits distributed over $D$ depth bins $\mathbf{b}=[b_1,\dots,b_D]^\top$ that are uniformly spaced in the log-depth domain.

The depth logits are then converted into a discrete probability distribution with a softmax operation, and the final predicted depth is computed as the expected value over these bins, yielding a depth map $\mathbf{D}^\ell_{g} \in \mathbb{R}^{H\times W}$ at stride $\ell$:
\begin{equation}
    \mathbf{D}_g^\ell(i,j) = \mathrm{softmax}(\tilde{\mathbf{D}}_g^\ell(i,j)) \cdot \mathbf{b}.
\end{equation}

To model depth uncertainty, a separate MLP predicts the log-variance for each pixel, producing a log-variance map $\mathbf{\Sigma}^\ell_{g} \in \mathbb{R}^{H \times W}$:
\begin{equation}
    \mathbf{\Sigma}_g^\ell = \Phi_{\mathrm{unc}}(\mathbf{F}_g^\ell),
\end{equation}
where a higher log-variance indicates greater depth uncertainty.

Both $\Phi_\mathrm{depth}$ and $\Phi_\mathrm{unc}$ share weights across all scales, allowing multi-scale structural cues to jointly guide depth estimation and produce a more reliable scene structure representation.

\subsubsection{Semantic Learning}

To develop stable semantics for cross-view localization, CROSS employs a \textit{semantic head} that predicts both semantic embeddings and reliability scores.
This dual output is motivated by the fact that transient or weakly grounded objects, such as cars and pedestrians, are often unreliable for establishing robust cross-view associations.

An MLP $\Phi_{\mathrm{sem}}$ first transforms both ground-view and satellite-view feature maps into semantic embedding maps:
\begin{equation}
    \mathbf{S}_g^\ell = \Phi_{\mathrm{sem}}(\mathbf{F}_g^\ell),\qquad
    \mathbf{S}_s^\ell = \Phi_{\mathrm{sem}}(\mathbf{F}_s^\ell),
\end{equation}
where $\mathbf{S}_g^\ell,\mathbf{S}_s^\ell \in \mathbb{R}^{C\times H \times W}$ for notational simplicity.
In practice, the ground-view image and satellite-view image may have different spatial resolutions.

A separate MLP $\Phi_{\mathrm{rel}}$ predicts reliability maps that estimate how trustworthy each pixel is for cross-view alignment:
\begin{equation}
    \mathbf{R}_g^\ell = \Phi_{\mathrm{rel}}(\mathbf{F}_g^\ell),\qquad
    \mathbf{R}_s^\ell = \Phi_{\mathrm{rel}}(\mathbf{F}_s^\ell),
\end{equation}
where $\mathbf{R}_g^\ell,\mathbf{R}_s^\ell \in \mathbb{R}^{H \times W}$.

Crucially, both $\Phi_{\mathrm{sem}}$ and $\Phi_\mathrm{rel}$ share weights across all scales and both viewpoints, encouraging CROSS to learn a consistent and generic cross-view semantic representation.

\subsection{3D-Grounded Alignment}

To address the lack of 3D grounding in existing formulations, we propose 3D-grounded alignment, which introduces a 2D-3D-2D pathway that requires the model to reason about scene structure.

Since ground-view images cover only a limited region compared with satellite-view images, we construct the spatial graph primarily from the ground-view image to reduce the introduction of irrelevant outliers.
Based on the multi-scale ground-view reliability maps $\mathbf{R}_g^\ell$ and semantic embedding maps $\mathbf{S}_g^\ell$, we extract a set of $N$ reliable nodes. 
For each $\ell \in \{1,2,4,8\}$, we apply non-maximum suppression to $\mathbf{R}_g^\ell$ and select the top-$N/4$ locations.
All selected locations are mapped to the original ground-view image resolution, yielding 2D node coordinates $\{\mathbf{x}_i\}_{i=1}^{N}$, where $\mathbf{x}_i\in\mathbb{R}^{2}$.

For each selected node $i$, we sample its semantic embedding $\mathbf{s}_i \in \mathbb{R}^C$, reliability score $r_i \in \mathbb{R}$, and predicted metric depth $d_i \in \mathbb{R}$:
\begin{equation}
    \mathbf{s}_i = \mathbf{S}^\ell_g(\hat{\mathbf{x}}_i), \quad r_i = \mathbf{R}^\ell_g(\hat{\mathbf{x}}_i), \quad d_i = \mathbf{D}^\ell_g(\hat{\mathbf{x}}_i), 
\end{equation}
where $\hat{\mathbf{x}}_i$ denotes the coordinates scaled to the corresponding feature resolution for sampling. 
This gives the ground-view spatial graph $\mathcal{G}_g$:
\begin{equation}
    \mathcal{G}_g = \{(\mathbf{x}_i, \mathbf{s}_i, r_i)\}_{i=1}^N.
\end{equation}

Given a ground-view camera pose hypothesis $\mathbf{H}=(\mathbf{R}, \mathbf{t})$, generated as described in Sec.~\ref{subsec:hypothesis-ranking}, each node is first back-projected into the 3D world coordinate frame using its predicted depth $d_i$:
\begin{equation}
    \mathbf{X}_i = \mathbf{R}(d_i \mathbf{K}_g^{-1}[\mathbf{x}_i^\top, 1]^\top)+\mathbf{t},
\end{equation}
where $\mathbf{K}_g$ denotes the ground-view camera intrinsic matrix.
The 3D point is then projected onto the satellite-view image plane:
\begin{equation}
    \mathbf{y}_i = \Pi_{\mathrm{sat}}(\mathbf{X}_i),
\end{equation}
where $\Pi_{\mathrm{sat}}$ denotes the projection mapping from world coordinates to satellite pixel coordinates.
This yields the 2D satellite-view graph $\mathcal{G}_s(\mathbf{H})$ induced by hypothesis $\mathbf{H}$:
\begin{equation}
    \mathcal{G}_s(\mathbf{H}) = \{(\mathbf{y}_i, \{\mathbf{s}_i^\ell\}, \{r_i^\ell\})\}_{i=1}^N,
\end{equation}
where the node at the coordinate $\mathbf{y}_i$ has multi-scale semantic embeddings $\{\mathbf{s}_i^\ell\}$ and reliability weights $\{r_i^\ell\}$ sampled from the multi-scale satellite-view semantic embedding maps $\{\mathbf{S}_s^\ell\}$ and reliability maps $\{\mathbf{R}_s^\ell\}$, and $\ell \in \{1,2,4,8\}$.

Through this 2D-3D-2D pathway, sparse cross-view correspondences between $\mathcal{G}_g$ and $\mathcal{G}_s(\mathbf{H})$ are explicitly grounded in the predicted 3D scene structure.
A pose hypothesis can therefore be evaluated according to whether the ground-view structure produces a spatially consistent alignment with the satellite view.

\subsection{Structure-Aware Matching}
To overcome the limitations of point-wise matching, we employ structure-aware matching to evaluate cross-view consistency by aggregating evidence over the holistic scene structure.
This encourages the model to learn semantic representations that remain stable across viewpoints.

Given a pose hypothesis $\mathbf{H}$, we evaluate its plausibility based on the ground-view spatial graph $\mathcal{G}_g$ and the hypothesis-induced satellite-view spatial graph $\mathcal{G}_s(\mathbf{H})$ obtained from 3D-grounded alignment.
For the $i$-th ground-view node, we compute the cosine similarity between its semantic embedding $\mathbf{s}_{i}$ and the corresponding satellite-view semantic embedding $\mathbf{s}_{i}^{\ell}$ at each scale $\ell$:
\begin{equation}
    c_i^\ell = \frac{\mathbf{s}_i \cdot \mathbf{s}_i^\ell}{\|\mathbf{s}_i\|_2 \, \|\mathbf{s}_i^\ell\|_2}.
\end{equation}

We also evaluate the reliability distribution of all correspondences by combining the predicted reliability score $r_i$ from the ground view and $r_i^\ell$ from the satellite view across all scales $\ell$:
\begin{equation}
    w_i^\ell = \frac{\mathrm{exp}(r_i + r_i^\ell)}{\sum_{i',\ell'} \mathrm{exp}(r_{i'} + r_{i'}^{\ell'})}.
\end{equation}
The weights are normalized over all $N$ nodes and all scales, allowing different correspondences to be emphasized at different resolutions.
For example, large objects such as buildings may be matched more reliably at coarse scales, whereas smaller objects such as trees may require finer-scale features.

Finally, the overall matching score for the given pose hypothesis $\mathbf{H}$ is computed by aggregating the similarity over the whole structure:
\begin{equation}
    m = \sum_i \sum_\ell w_i^\ell c_i^\ell.
\end{equation}

Notably, structure-aware matching evaluates each hypothesis based on the holistic spatial consistency rather than isolated point-wise similarities.
Under an incorrect hypothesis, a few isolated nodes, such as road regions, may coincidentally project onto visually similar areas and yield high local similarities.
However, the overall matching score $m$ remains low when the majority of the structure is misaligned.
This avoids forcing the model to exclusively discriminate semantically consistent objects across viewpoints and instead encourages semantic representations to remain stable under extreme viewpoint changes.

\subsection{Hypothesis Ranking}
\label{subsec:hypothesis-ranking}

Structure-aware matching enables CROSS to evaluate a pose hypothesis $\mathbf{H}$ by its spatial consistency.
Instead of optimizing an absolute objective such as direct pose regression, CROSS learns with a relative hypothesis ranking objective.
This objective unifies fully and weakly supervised training and encourages transferable geometry for unseen transformations.

\subsubsection{Fully Supervised Training}

In the fully supervised setting, where the ground-truth 3-DoF camera pose $(x^*, y^*, \theta^*)$ is available, CROSS learns to rank the ground-truth pose hypothesis $\mathbf{H}^{*}$ above alternative negative hypotheses.

\begin{figure}[!t]
    \centering
    \includegraphics[width=0.95\columnwidth]{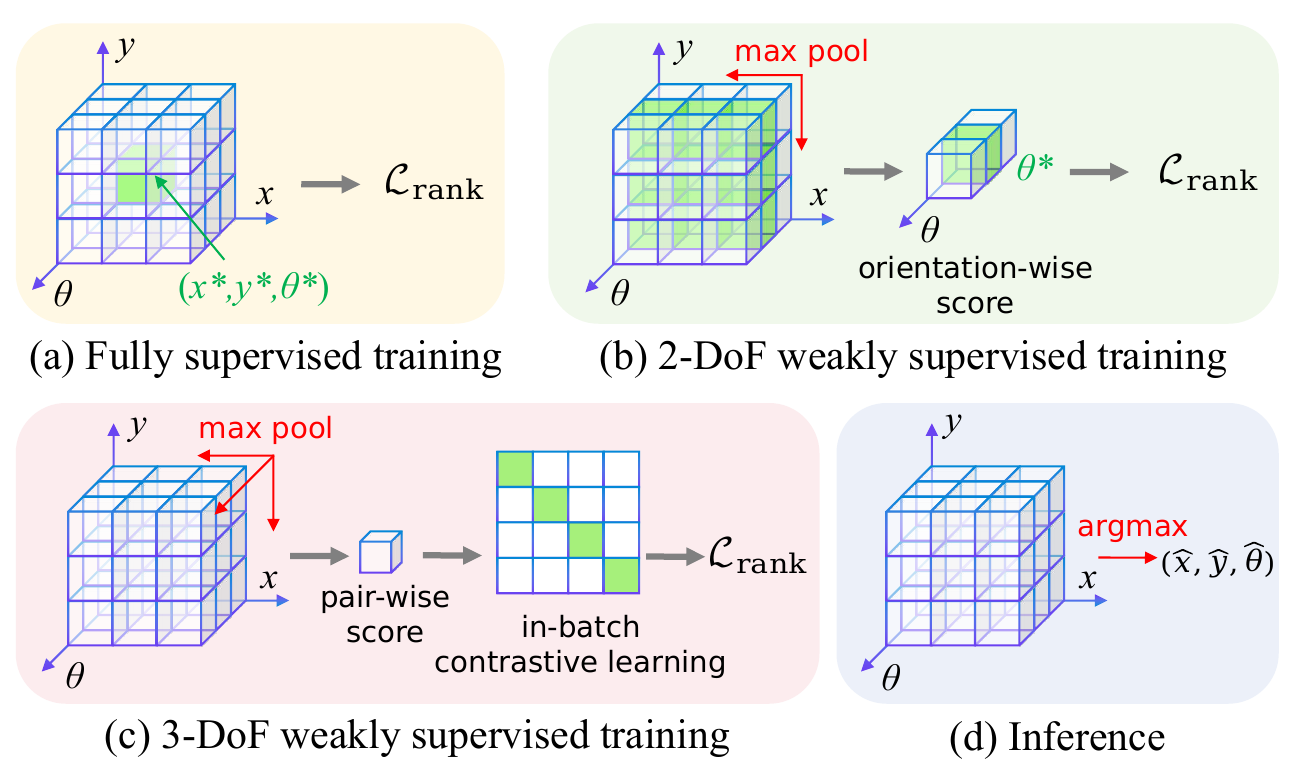}
    \caption{Illustration of hypothesis ranking for fully and weakly supervised training and inference.}
    \label{fig:hypothesis-ranking}
\end{figure}

Specifically, as shown in Fig.~\ref{fig:hypothesis-ranking}(a), we generate $K$ negative hypotheses $\{\mathbf{H}_k\}_{k=1}^K$ by applying perturbations to $\mathbf{H}^{*}$.
Structure-aware matching produces a score $m_k$ for each negative hypothesis $\mathbf{H}_k$, and $m^*$ denotes the score of the ground-truth hypothesis.
We adopt a Negative Log-Likelihood (NLL) loss~\cite{oord2018cpc} to prioritize the ground-truth hypothesis:
\begin{equation}
    \mathcal{L}_\mathrm{rank} = -\mathrm{log}\left(\frac{\mathrm{exp}(m^*/\tau)}{\mathrm{exp}(m^*/\tau) + \sum_{k=1}^K\mathrm{exp}(m_k/\tau)}\right),
\end{equation}
where $\tau$ is a temperature hyperparameter.

\subsubsection{Weakly Supervised Training}

In the weakly supervised setting, the ground-truth label for the 3-DoF camera pose is unavailable.
We consider two weak supervision settings: \textbf{2-DoF weak supervision}, where the horizontal translation $(x^*,y^*)$ is unavailable while the orientation angle $\theta^*$ is provided following the setting adopted by most previous weakly supervised cross-view localization methods, and the more challenging \textbf{3-DoF weak supervision} setting, where accurate labels for both translation and orientation are unavailable.

\textbf{2-DoF Weak Supervision}: When only the orientation label $\theta^*$ is available, CROSS learns to rank hypotheses with the correct orientation above those with incorrect orientations, while marginalizing over the unknown horizontal translation.
Specifically, as shown in Fig.~\ref{fig:hypothesis-ranking}(b), we sample $K_{\mathrm{ori}}$ negative orientation candidates and generate $K_{\mathrm{trans}}$ translation hypotheses for each orientation by perturbing the initial translation $(\tilde{x}, \tilde{y})$.
Together with the ground-truth orientation $\theta^*$, this yields $K_{\mathrm{trans}}$ potential positive hypotheses and $K_{\mathrm{ori}} \times K_{\mathrm{trans}}$ negative hypotheses.
CROSS computes a matching score for each hypothesis.
Since the translation label is unavailable, we max-pool the scores over hypotheses sharing the same orientation, producing one score $m^*$ for the ground-truth orientation and $K_{\mathrm{ori}}$ scores $\{m_k\}_{k=1}^{K_{\mathrm{ori}}}$ for negative orientations.
The ranking loss is then defined as:
\begin{equation}
    \mathcal{L}_{\mathrm{rank}}
    =
    -\log
    \left(
    \frac{\exp(m^*/\tau)}
    {\exp(m^*/\tau) + \sum_{k=1}^{K_{\mathrm{ori}}}\exp(m_k/\tau)}
    \right).
\end{equation}

\textbf{3-DoF Weak Supervision}: In this more challenging setting, where only a noisy initial pose $(\tilde{x}, \tilde{y}, \tilde{\theta})$ is provided, CROSS learns to rank the most plausible pose within an aligned image pair above those from misaligned pairs.

Given a batch of $B$ aligned ground-view and satellite-view images with noisy initial poses, denoted as $\{(\mathbf{I}_{i,g}, \mathbf{I}_{i,s}, (\tilde{x}_i, \tilde{y}_i, \tilde{\theta}_i))\}_{i=1}^B$, we perturb each noisy pose $(\tilde{x}_i, \tilde{y}_i, \tilde{\theta}_i)$ to cover the surrounding search region.
This yields $K$ pose hypotheses $\{\mathbf{H}_{i,k}\}_{k=1}^K$ for pair $i$.

As illustrated in Fig.~\ref{fig:hypothesis-ranking}(c), CROSS then evaluates matching scores across both positive and negative pairs.
For every combination $(\mathbf{I}_{i,g}, \mathbf{I}_{j,s}, \{\mathbf{H}_{j,k}\}_{k=1}^K)$, where $i=j$ denotes an aligned pair and $i \neq j$ a misaligned pair, the matching scores of the $K$ hypotheses are computed. 
These scores are max-pooled into a single score $m_{i,j}$, which represents the most plausible pose hypothesis for the image pair $(i,j)$.
An NLL loss is then applied to train the model to rank the aligned pair above the misaligned pairs:
\begin{equation}
    \mathcal{L}_\mathrm{rank} = \sum_{i=1}^B -\mathrm{log}\left(\frac{\mathrm{exp}(m_{i,i} / \tau)}{\sum_{j=1}^B \mathrm{exp}(m_{i,j} / \tau)}\right).
\end{equation}
This allows CROSS to distinguish spatially consistent hypotheses from misaligned alternatives even without precise pose supervision.

\subsubsection{Overall Loss}
Although foundation models such as Depth Anything 3 fail to predict reliable and scale-consistent structure across extremely different viewpoints, they still provide strong priors for relative structure.
We therefore introduce a monocular-guided loss to distill this prior and refine a scale-consistent structure across viewpoints.

For a ground-view image $\mathbf{I}_g$, CROSS predicts multi-scale monocular metric-scale depth maps $\mathbf{D}_g^\ell$ and log-variance maps $\mathbf{\Sigma}_g^\ell$.
Based on the monocular depth map $\mathbf{D}_g$ predicted by foundation models~\cite{lin2025da3,lin2026dap} with arbitrary scale and interpolated to the same resolution as $\mathbf{D}_g^\ell$, we compute the confidence $c_i$ and log residual $g_i$ for each pixel $i$:
\begin{equation}
    c_i = \mathrm{exp}(-\sigma_i),~~ g_i = \log(\hat{d}_i) - \log(d_i),
\end{equation}
where $\sigma_i \in \mathbf{\Sigma}_g^\ell$, $\hat{d}_i \in \mathbf{D}_g^\ell$, and $d_i \in \mathbf{D}_g$ represent the predicted log-variance, predicted metric-scale depth and the monocular depth at the $i$-th pixel.
The global scale shift $\mu$ is then computed as:
\begin{equation}
    \mu = \frac{\sum_i c_i g_i}{\sum_i c_i},
\end{equation}
and the monocular-guided loss is defined as the combination of a scale-invariant error and an uncertainty regularization penalty:
\begin{equation}
    \mathcal{L}_\mathrm{mono} = \sum_i c_i (g_i - \mu)^2 + \sum_i \sigma_i.
\end{equation}
The overall training objective integrates this monocular guidance with the ranking loss:
\begin{equation}
    \mathcal{L} = \mathcal{L}_\mathrm{rank} + \lambda \mathcal{L}_\mathrm{mono},
\end{equation}
where we empirically set the weighting hyperparameter $\lambda = 0.1$.

\subsubsection{Inference}

During inference, as shown in Fig.~\ref{fig:hypothesis-ranking}(d), given an input ground-view image $\mathbf{I}_g$, a satellite-view image $\mathbf{I}_s$, and an initial 3-DoF camera pose $(\tilde{x}, \tilde{y}, \tilde{\theta})$, we first generate pose hypotheses by perturbing the initial pose to cover the surrounding search region. 
This yields $K$ candidate pose hypotheses $\{\mathbf{H}_k\}_{k=1}^K$.
CROSS then evaluates the spatial consistency of each hypothesis, computing a corresponding matching score $m_k$. 
The final predicted 3-DoF camera pose is derived from the most plausible hypothesis:
\begin{equation}
    \hat{\mathbf{H}} = \underset{\mathbf{H}_k}{\arg\max} \; m_k, \quad k \in \{1, 2, \dots, K\},
\end{equation}
where the final pose prediction $(\hat{x}, \hat{y}, \hat{\theta})$ is directly extracted from the optimal hypothesis $\hat{\mathbf{H}}$.

We also conduct coarse-to-fine search which significantly improves inference efficiency as described in Sec.~\ref{subsec:efficiency}.

\section{Experiments}
In this section, we conduct extensive experiments on two cross-view localization datasets to validate the effectiveness of our method in both localization accuracy and semantic, structural, and geometric learning.

\subsection{Datasets, Metrics, and Implementation Details}

\subsubsection{Datasets}

We evaluate the proposed method on two widely used cross-view localization benchmarks: KITTI~\cite{geiger2012kitti} and VIGOR~\cite{zhu2021vigor}. These datasets cover complementary challenges, including limited ground-view fields of view, large viewpoint gaps, and substantial spatial displacement between ground and satellite views.

\textbf{KITTI} is a standard benchmark collected primarily in urban driving scenarios. The ground-view queries are front-facing camera images, which naturally have a limited field of view (FoV). The corresponding satellite maps cover a $100~\text{m} \times 100~\text{m}$ region and are centered according to GPS coordinates aligned with the ground-view camera~\cite{shi2022beyond}.
The dataset is evaluated under two spatial splits: \textit{same-area}, where training and testing samples are drawn from the same driving trajectories, and \textit{cross-area}, where testing is performed on geographically disjoint trajectories.
Following previous works~\cite{xia2026loc2,xia2025fg2}, we conduct experiments with an initial translation noise of $\pm 20~\text{m}$ and two levels of initial orientation noise, i.e., $\pm 10^\circ$ and $\pm 180^\circ$.

\textbf{VIGOR} is a large-scale and realistic cross-view localization benchmark covering four major US cities: New York, Seattle, San Francisco, and Chicago.
The ground-view queries are panoramic images, and their true locations are randomly displaced within the central $1/4$ region of the corresponding satellite-view images.
Similar to KITTI, VIGOR provides two evaluation splits: \textit{same-area}, where training and testing data share geographic regions, and \textit{cross-area}, where the model is evaluated on unseen cities.
Following previous works~\cite{xia2025fg2,xia2026loc2,tong2025geodistill}, we evaluate all models under initial orientation noise of $0^\circ$ and $\pm180^\circ$ in the fully supervised setting, and $\pm45^\circ$ and $\pm180^\circ$ in the weakly supervised setting.

\subsubsection{Metrics}

\begin{table*}[t]
\centering
\caption{Fully supervised cross-view localization test results on the KITTI dataset. Results are reported under initial orientation noise of $\pm10^\circ$ and $\pm180^\circ$. The \protect\colorbox{bestcolor}{best performance} and the \protect\colorbox{secondcolor}{second-best performance} are highlighted.}
\begin{tabular}{clcccccccc}
\toprule
& & \multicolumn{4}{c}{$\pm 10^\circ$} & \multicolumn{4}{c}{$\pm 180^\circ$} \\
Area & Methods 
& \multicolumn{2}{c}{$\downarrow$ Localization (m)} 
& \multicolumn{2}{c}{$\downarrow$ Orientation ($^\circ$)} 
& \multicolumn{2}{c}{$\downarrow$ Localization (m)} 
& \multicolumn{2}{c}{$\downarrow$ Orientation ($^\circ$)} \\
& & Mean & Median & Mean & Median & Mean & Median & Mean & Median \\
\midrule
\multirow{8}{*}{\rotatebox{90}{Cross-area}} 
& SliceMatch~~(CVPR'23) 
& - & - & - & - 
& 14.85 & 11.85 & 23.64 & 7.96 \\

& CCVPE~~(TPAMI'24) 
& 9.16 & 3.33 & 1.55 & \best{0.84} 
& 13.94 & 10.98 & 77.84 & 63.84 \\

& HC-Net~~(NeurIPS'23) 
& 8.47 & 4.57 & 3.22 & 1.63 
& - & - & - & - \\

& DenseFlow~~(NeurIPS'23) 
& 7.97 & 3.52 & 2.17 & 1.21 
& - & - & - & - \\

& FG$^2$~~(CVPR'25) 
& 7.31 & 4.15 & 3.62 & 2.37 
& - & - & - & - \\

& Loc$^2$~~(ICLR'26) 
& 5.60 & 3.01 & 3.32 & 2.12 
& 11.71 & 9.11 & 55.18 & 33.23 \\

& CROSS-DINOv2~~(Ours) 
& \second{3.24} & \second{2.00} & \best{1.32} & \second{0.96} 
& \second{3.50} & \second{2.11} & \best{4.68} & \best{1.20} \\

& CROSS-DINOv3~~(Ours) 
& \best{3.11} & \best{1.82} & \second{1.37} & 0.99 
& \best{3.37} & \best{1.89} & \second{5.07} & \second{1.24} \\

\midrule
\multirow{8}{*}{\rotatebox{90}{Same-area}} 
& SliceMatch~~(CVPR'23) 
& - & - & - & - 
& 7.96 & 4.39 & 4.12 & 3.65 \\

& CCVPE~~(TPAMI'24) 
& 1.22 & 0.62 & 0.67 & 0.54 
& 6.88 & 3.47 & 15.01 & 6.12 \\

& HC-Net~~(NeurIPS'23) 
& \second{0.80} & \second{0.50} & \best{0.45} & \second{0.33} 
& - & - & - & - \\

& DenseFlow~~(NeurIPS'23) 
& 1.48 & \best{0.47} & \second{0.49} & \best{0.30} 
& - & - & - & - \\

& FG$^2$~~(CVPR'25) 
& \best{0.75} & 0.51 & 0.93 & 0.66 
& - & - & - & - \\

& Loc$^2$~~(ICLR'26) 
& 1.13 & 0.77 & 1.97 & 1.43 
& \best{1.85} & \best{1.31} & 9.70 & 6.17 \\

& CROSS-DINOv2~~(Ours) 
& 2.86 & 1.87 & 1.20 & 0.81 
& 2.98 & 1.96 & \second{2.70} & \best{1.07} \\

& CROSS-DINOv3~~(Ours) 
& 2.28 & 1.43 & 1.17 & 0.82 
& \second{2.40} & \second{1.52} & \best{2.05} & \second{1.08} \\
\bottomrule
\end{tabular}
\label{tab:kitti-full}
\end{table*}

\begin{table*}[t]
\centering
\caption{Fully supervised cross-view localization test results on the VIGOR dataset. 
Results are reported under known and unknown orientation. 
The \protect\colorbox{bestcolor}{best performance} and the \protect\colorbox{secondcolor}{second-best performance} are highlighted.}
\begin{tabular}{clcccccccc}
\toprule
& & \multicolumn{4}{c}{Unknown} & \multicolumn{4}{c}{Known} \\
\cmidrule(lr){3-6} \cmidrule(lr){7-10}
Area & Methods 
& \multicolumn{2}{c}{$\downarrow$ Localization (m)} 
& \multicolumn{2}{c}{$\downarrow$ Orientation ($^\circ$)} 
& \multicolumn{2}{c}{$\downarrow$ Localization (m)} 
& \multicolumn{2}{c}{$\downarrow$ Orientation ($^\circ$)} \\
& & Mean & Median & Mean & Median & Mean & Median & Mean & Median \\
\midrule
\multirow{9}{*}{\rotatebox{90}{Cross-area}} 
& SliceMatch~~(CVPR'23) 
& 7.22 & 3.31 & 25.97 & 4.51 
& 5.53 & 2.55 & - & - \\

& CCVPE~~(TPAMI'24) 
& 5.41 & 1.89 & 27.78 & 13.58 
& 4.97 & 1.68 & - & - \\

& GGCVT~~(ECCV'24) 
& - & - & - & - 
& 5.16 & 1.40 & - & - \\

& DenseFlow~~(NeurIPS'23) 
& 7.67 & 3.67 & 17.63 & 2.94 
& 5.01 & 2.42 & - & - \\

& HC-Net~~(NeurIPS'23) 
& - & - & - & - 
& 3.35 & 1.59 & - & - \\

& FG$^2$~~(CVPR'25) 
& 10.02 & 8.14 & 31.41 & 5.45 
& 2.41 & 1.37 & - & - \\

& Loc$^2$~~(ICLR'26) 
& 4.23 & 2.09 & 11.67 & 2.21 
& 3.43 & 1.90 & - & - \\

& CROSS-DINOv2~~(Ours) 
& \second{2.25} & \second{1.22} & \best{5.08} & \second{1.17} 
& \second{2.00} & \second{1.19} & - & - \\

& CROSS-DINOv3~~(Ours) 
& \best{2.24} & \best{1.15} & \second{5.16} & \best{1.16} 
& \best{1.98} & \best{1.12} & - & - \\

\midrule
\multirow{9}{*}{\rotatebox{90}{Same-area}} 
& SliceMatch~~(CVPR'23) 
& 6.49 & 3.13 & 25.46 & 4.71 
& 5.18 & 2.58 & - & - \\

& CCVPE~~(TPAMI'24) 
& 3.74 & 1.42 & 12.83 & 6.62 
& 3.60 & 1.36 & - & - \\

& GGCVT~~(ECCV'24) 
& - & - & - & - 
& 4.12 & 1.34 & - & - \\

& DenseFlow~~(NeurIPS'23) 
& 4.97 & 1.90 & 11.20 & 1.59 
& 3.03 & \best{0.97} & - & - \\

& HC-Net~~(NeurIPS'23) 
& - & - & - & - 
& 2.65 & 1.17 & - & - \\

& FG$^2$~~(CVPR'25) 
& 8.95 & 7.32 & 15.02 & 2.94 
& \best{1.95} & \second{1.08} & - & - \\

& Loc$^2$~~(ICLR'26) 
& 3.94 & 1.78 & 9.54 & 2.00 
& 3.06 & 1.59 & - & - \\

& CROSS-DINOv2~~(Ours) 
& \best{2.34} & \second{1.24} & \best{4.28} & \second{1.20} 
& \second{2.13} & 1.21 & - & - \\

& CROSS-DINOv3~~(Ours) 
& \second{2.36} & \best{1.20} & \second{4.68} & \best{1.19} 
& \second{2.13} & 1.18 & - & - \\
\bottomrule
\end{tabular}
\label{tab:vigor-full}
\end{table*}

We evaluate cross-view localization performance using the mean and median localization errors in meters, as well as the mean and median orientation errors in degrees.

\subsubsection{Implementation Details}

We evaluate our model with both DINOv2~\cite{oquab2024dinov2} and DINOv3~\cite{simeoni2025dinov3} as frozen backbone encoders. This setting enables a fair comparison with existing baselines, which predominantly adopt DINOv2, while also demonstrating the performance gains brought by a stronger visual foundation model.
For spatial representation learning, the DPT decoder outputs feature maps with channel dimension $C=64$, and the semantic head further reduces the feature dimension to 8.
Meanwhile, we use $D=64$ depth bins ranging from $1~\text{m}$ to $100~\text{m}$ for structural learning.
For 3D-grounded alignment, we construct a spatial graph with $N=256$ nodes and initialize the temperature parameter $\tau$ as 0.07.
We use monocular depth maps predicted by Depth Anything 3~\cite{lin2025da3} and DAP~\cite{lin2026dap} for KITTI and VIGOR, respectively, to provide monocular guidance during training.
During training, we generate up to $K = 17^3 = 4913$ candidate pose hypotheses for ranking.
During inference, $K$ can be further increased to achieve more precise localization.
The model is trained for 20 epochs using the AdamW optimizer~\cite{loshchilov2019adamw}, with a weight decay of 0.01 and a batch size of $B=8$.
The learning rate is linearly warmed up to $1 \times 10^{-4}$ over the first 500 steps and then decayed with a cosine annealing schedule.
All experiments are conducted on a single NVIDIA RTX 6000 Ada Generation GPU.

\subsection{Cross-View Localization Comparison}
\label{subsec:localization-results}
We compare the proposed CROSS with a broad range of cross-view localization methods under both fully and weakly supervised settings.

\subsubsection{Fully Supervised Comparison}

In the fully supervised setting, accurate 3-DoF pose labels are available during training.
We compare CROSS with representative methods from different localization paradigms, including global matching methods such as SliceMatch~\cite{lentsch2023slicematch} and CCVPE~\cite{xia2024ccvpe}, dense matching methods such as HC-Net~\cite{wang2023hcnet} and DenseFlow~\cite{song2023denseflow}, and recent sparse matching methods such as FG$^2$~\cite{xia2025fg2} and Loc$^2$~\cite{xia2026loc2}.

As shown in Tab.~\ref{tab:kitti-full} and Tab.~\ref{tab:vigor-full}, global matching methods such as SliceMatch and CCVPE generally achieve inferior performance, as they compress the ground-view image into a single global descriptor and inevitably discard fine-grained spatial details.
Recent sparse matching methods, including FG$^2$ and Loc$^2$, usually outperform dense matching methods such as DenseFlow and HC-Net.
This is because sparse matching allows the model to focus on discriminative landmarks for cross-view localization, rather than relying on dense correspondences that can be ambiguous under extreme viewpoint changes.
Nevertheless, CROSS consistently achieves stronger performance through 3D-grounded alignment, structure-aware matching, and hypothesis ranking.

On the KITTI dataset, as reported in Tab.~\ref{tab:kitti-full}, CROSS achieves particularly strong performance in the cross-area setting, which better reflects the generalization ability of different methods.
Under small orientation noise ($\pm10^\circ$), CROSS substantially reduces the localization error compared with previous methods.
For example, CROSS-DINOv2 obtains a mean/median localization error of $3.24/2.00$ m, clearly outperforming Loc$^2$ with $5.60/3.01$ m.
Under large orientation noise ($\pm180^\circ$), the advantage becomes even more significant.
CROSS-DINOv2 reduces the mean localization error from $11.71$ m of Loc$^2$ to $3.50$ m, corresponding to an approximately $70\%$ relative reduction, while also achieving much lower orientation errors.
These results demonstrate that CROSS is more robust to severe orientation ambiguity and generalizes more effectively to unseen areas.
Moreover, while CROSS-DINOv2 already outperforms most baselines under a fair backbone setting, CROSS-DINOv3 further improves localization accuracy, showing the benefit of stronger visual foundation representations.

In the KITTI same-area setting, CROSS remains competitive, especially under large orientation noise, where it achieves the best orientation accuracy.
However, its localization performance is inferior to several baselines in this setting, which should be interpreted with caution.
In the same-area split, training and testing frames are sampled from the same driving trajectories and therefore share substantial overlap in scene appearance and spatial layout.
Such overlap may allow methods that directly optimize absolute localization targets to overfit to trajectory-specific cues and achieve high accuracy by memorizing nearby frames observed during training.
This tendency is supported by their significant performance gap between the same-area and cross-area settings.
In contrast, the smaller gap achieved by CROSS suggests that it learns a more transferable localization mechanism, further validating the effectiveness of the proposed hypothesis ranking strategy.

On the VIGOR dataset, as shown in Tab.~\ref{tab:vigor-full}, CROSS consistently outperforms existing baselines across both cross-area and same-area settings.
In the cross-area setting with unknown orientation, CROSS-DINOv2 achieves the best localization accuracy with a mean/median error of $2.25/1.22$ m, substantially improving over previous methods such as Loc$^2$ ($4.23/2.09$ m) and CCVPE ($5.41/1.89$ m).
Meanwhile, CROSS also significantly reduces the orientation error, achieving the best mean orientation error of $5.08^\circ$.
When orientation is known, CROSS remains superior, obtaining a mean/median localization error of $2.00/1.19$ m and outperforming both dense matching methods and sparse matching baselines.
Similar trends are observed in the same-area setting, where CROSS achieves the best localization and orientation performance under unknown orientation and remains competitive when orientation is known.
These results demonstrate that CROSS enables accurate cross-view localization while maintaining strong robustness to orientation uncertainty and reliable generalization across different geographic regions.

\subsubsection{Weakly Supervised Comparison}

\begin{table*}[t]
\centering
\caption{Weakly supervised cross-view localization test results on the KITTI dataset. Results are reported under initial orientation noise of $\pm10^\circ$ and $\pm180^\circ$. The \protect\colorbox{bestcolor}{best performance} is highlighted. All models use DINOv2 as the backbone.}
\begin{tabular}{clcccccccc}
\toprule
& & \multicolumn{4}{c}{$\pm 10^\circ$} & \multicolumn{4}{c}{$\pm 180^\circ$} \\
Area & Methods 
& \multicolumn{2}{c}{$\downarrow$ Localization (m)} 
& \multicolumn{2}{c}{$\downarrow$ Orientation ($^\circ$)} 
& \multicolumn{2}{c}{$\downarrow$ Localization (m)} 
& \multicolumn{2}{c}{$\downarrow$ Orientation ($^\circ$)} \\
& & Mean & Median & Mean & Median & Mean & Median & Mean & Median \\
\midrule
\multirow{6}{*}{\rotatebox{90}{Cross-area}} 
& \multicolumn{9}{l}{\textbf{\textit{2-DoF weakly-supervised training}}} \\
& G2SWeakly~~(ECCV'24) 
& 12.61 & 11.64 & N/A & N/A 
& 17.16 & 16.32 & N/A & N/A \\

& GeoDistill~~(ICCV'25) 
& 11.85 & 11.17 & N/A & N/A 
& 16.78 & 15.98 & N/A & N/A \\

& CROSS~~(Ours) 
& \best{7.04} & \best{4.02} & \best{1.76} & \best{1.14}
& \best{7.93} & \best{4.50} & \best{10.59} & \best{1.49} \\

& \multicolumn{9}{l}{\textbf{\textit{3-DoF weakly-supervised training}}} \\
& CROSS~~(Ours) 
& \best{9.01} & \best{6.01} & \best{3.05} & \best{1.99}
& \best{11.02} & \best{7.61} & \best{25.76} & \best{3.06} \\

\midrule
\multirow{6}{*}{\rotatebox{90}{Same-area}} 
& \multicolumn{9}{l}{\textbf{\textit{2-DoF weakly-supervised training}}} \\
& G2SWeakly~~(ECCV'24) 
& 11.68 & 10.96 & N/A & N/A 
& 16.86 & 15.99 & N/A & N/A \\

& GeoDistill~~(ICCV'25) 
& 11.52 & 10.91 & N/A & N/A 
& 16.50 & 15.42 & N/A & N/A \\

& CROSS~~(Ours) 
& \best{6.84} & \best{3.96} & \best{1.81} & \best{1.26}
& \best{7.19} & \best{4.14} & \best{3.77} & \best{1.55} \\

& \multicolumn{9}{l}{\textbf{\textit{3-DoF weakly-supervised training}}} \\
& CROSS~~(Ours) 
& \best{8.33} & \best{5.29} & \best{3.01} & \best{2.04}
& \best{9.57} & \best{6.14} & \best{20.33} & \best{2.80} \\
\bottomrule
\end{tabular}
\label{tab:kitti-weak}
\end{table*}

\begin{table*}[!h]
\centering
\caption{Weakly-supervised cross-view localization test results on the VIGOR dataset. Results are reported under initial orientation noise of $\pm45^\circ$ and $\pm180^\circ$. The \protect\colorbox{bestcolor}{best performance} is highlighted. All models use DINOv2 as the backbone.}
\begin{tabular}{clcccccccc}
\toprule
& & \multicolumn{4}{c}{$\pm 45^\circ$} & \multicolumn{4}{c}{$\pm 180^\circ$} \\
Area & Methods 
& \multicolumn{2}{c}{$\downarrow$ Localization (m)} 
& \multicolumn{2}{c}{$\downarrow$ Orientation ($^\circ$)} 
& \multicolumn{2}{c}{$\downarrow$ Localization (m)} 
& \multicolumn{2}{c}{$\downarrow$ Orientation ($^\circ$)} \\
& & Mean & Median & Mean & Median & Mean & Median & Mean & Median \\
\midrule
\multirow{6}{*}{\rotatebox{90}{Cross-area}} 
& \multicolumn{9}{l}{\textbf{\textit{2-DoF weakly-supervised training}}} \\
& G2SWeakly~~(ECCV'24) 
& 5.24 & 2.88 & 4.79 & 1.58 
& 15.27 & 13.70 & 90.26 & 89.99 \\

& GeoDistill~~(ICCV'25) 
& 4.50 & 2.62 & 4.79 & 1.58
& 15.58 & 13.89 & 90.26 & 89.99 \\

& CROSS~~(Ours) 
& \best{2.98} & \best{1.61} & \best{2.13} & \best{1.14}
& \best{3.19} & \best{1.63} & \best{6.94} & \best{1.31} \\

& \multicolumn{9}{l}{\textbf{\textit{3-DoF weakly-supervised training}}} \\
& CROSS~~(Ours) 
& \best{3.88} & \best{2.11} & \best{3.78} & \best{1.86}
& \best{4.37} & \best{2.15} & \best{11.43} & \best{1.97} \\

\midrule
\multirow{6}{*}{\rotatebox{90}{Same-area}} 
& \multicolumn{9}{l}{\textbf{\textit{2-DoF weakly-supervised training}}} \\
& G2SWeakly~~(ECCV'24) 
& 5.66 & 3.11 & 3.68 & 1.29 
& 15.39 & 13.83 & 90.05 & 90.07 \\

& GeoDistill~~(ICCV'25) 
& 4.57 & 2.74 & 3.68 & 1.29 
& 14.69 & 12.88 & 90.05 & 90.07 \\

& CROSS~~(Ours) 
& \best{3.09} & \best{1.78} & \best{2.19} & \best{1.21}
& \best{3.24} & \best{1.80} & \best{5.62} & \best{1.35} \\

& \multicolumn{9}{l}{\textbf{\textit{3-DoF weakly-supervised training}}} \\
& CROSS~~(Ours) 
& \best{4.29} & \best{2.62} & \best{4.09} & \best{1.97}
& \best{4.63} & \best{2.66} & \best{11.80} & \best{2.07} \\
\bottomrule
\end{tabular}
\label{tab:vigor-weak}
\end{table*}

Tab.~\ref{tab:kitti-weak} and Tab.~\ref{tab:vigor-weak} report weakly supervised cross-view localization results under noisy pose labels.
In the standard 2-DoF weakly supervised setting, accurate position labels $(x^*,y^*)$ are unavailable during training, while the orientation label $\theta^*$ remains accurate.
Existing methods such as G2SWeakly~\cite{shi2024g2sweakly} and GeoDistill~\cite{tong2025geodistill} typically handle translation and orientation separately.
On KITTI, the baselines do not explicitly estimate orientation, and their localization performance remains competitive only when the initial orientation noise is small.
When the initial orientation noise increases from $\pm10^\circ$ to $\pm180^\circ$, their localization errors increase substantially, indicating that inaccurate orientation can severely degrade translation estimation.
On VIGOR, the baselines rely on an independent CNN-based orientation estimator, which performs well under moderate initial orientation noise of $\pm45^\circ$ and achieves a mean orientation error of about $4^\circ$.
However, this estimator breaks down under an initial orientation noise of $\pm180^\circ$, producing nearly random orientation errors around $90^\circ$ and causing the localization error to increase from about $5$ m to about $15$ m.

In contrast, CROSS does not decouple orientation from translation.
Instead, it represents $(x,y,\theta)$ as a joint pose hypothesis and ranks candidate hypotheses according to their cross-view consistency.
This coupled formulation allows CROSS to consistently outperform previous methods in the 2-DoF setting, especially under large orientation noise.
On KITTI, CROSS reduces the mean localization error under $\pm180^\circ$ orientation noise to $7.93$ m and $7.19$ m on the cross-area and same-area splits, respectively.
On VIGOR, the improvement is even more significant.
Under $\pm180^\circ$ orientation noise, CROSS reduces the mean localization error from about $15$ m to around $3$ m, while maintaining accurate orientation estimation with mean errors of $6.94^\circ$ and $5.62^\circ$ on the cross-area and same-area splits, respectively.
These results demonstrate that the hypothesis ranking objective enables the model to jointly reason about translation and rotation, which is crucial for robust weakly supervised localization.

More importantly, CROSS naturally extends to the more challenging 3-DoF weakly supervised setting, where all pose labels $(x^*,y^*,\theta^*)$ are unavailable during training.
This setting cannot be directly accommodated by previous weakly supervised methods, as they rely on accurate orientation labels to train a separate orientation estimator.
Even under this substantially weaker supervision, CROSS remains robust and still outperforms the 2-DoF weakly supervised baselines across both datasets.
For example, under large orientation noise, CROSS achieves mean localization errors of $11.02$ m and $9.57$ m on KITTI, and $4.37$ m and $4.63$ m on VIGOR, consistently surpassing previous 2-DoF baselines.
These results show that the proposed hypothesis ranking objective can simultaneously tolerate noisy translation and orientation labels, making CROSS better suited to realistic weakly supervised scenarios where accurate 3-DoF pose annotations are difficult to obtain.

\subsubsection{Visualization of Localization and Matching}

\begin{figure*}
    \centering
    \includegraphics[width=0.95\linewidth]{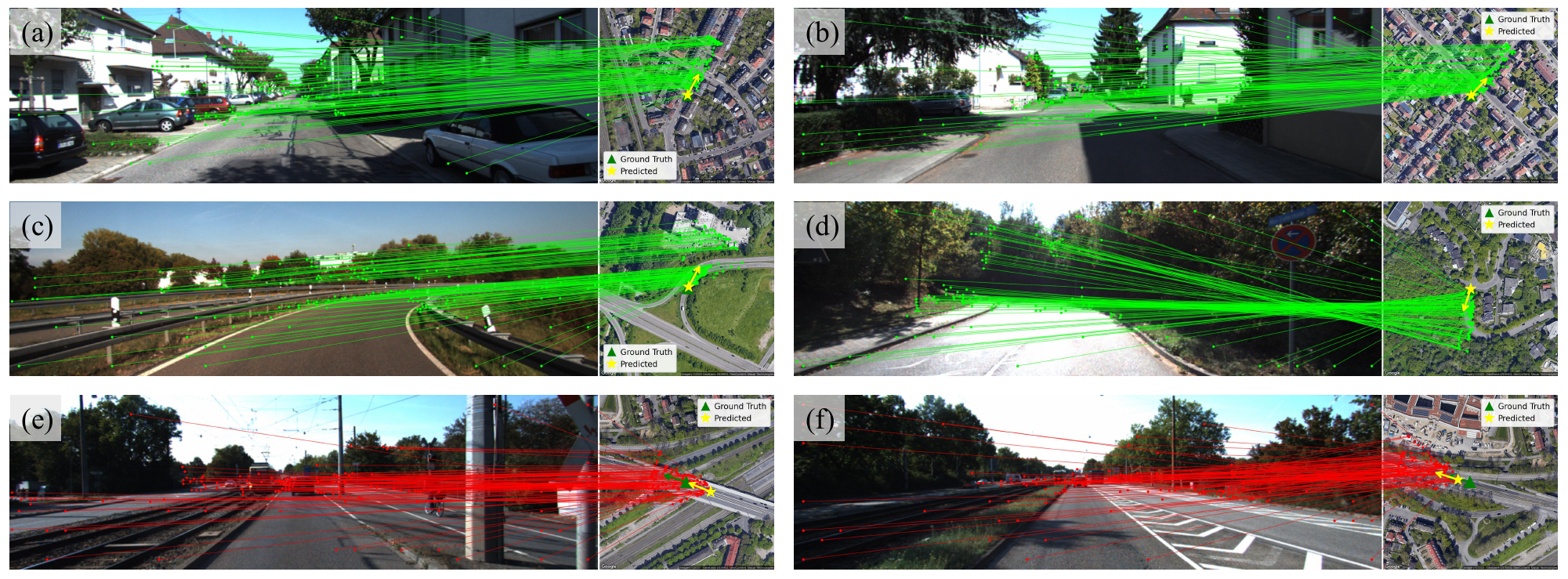}
    \caption{Visualization of cross-view localization and matching results on the KITTI dataset under unknown orientation. Green and red lines denote matching results in successful and failed cases, respectively. The satellite inset compares the ground-truth pose and the predicted pose.}
    \label{fig:kitti_loc}
\end{figure*}

\begin{figure*}
    \centering
    \includegraphics[width=0.95\linewidth]{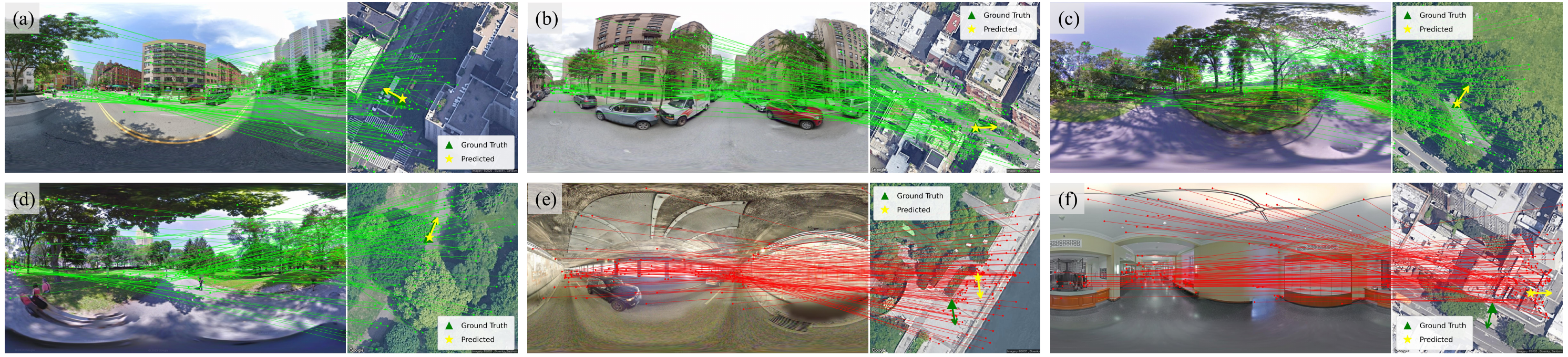}
    \caption{Visualization of cross-view localization and matching results on the VIGOR dataset under unknown orientation. Green and red lines denote matching results in successful and failed cases, respectively. The satellite inset compares the ground-truth pose and the predicted pose.}
    \label{fig:vigor_loc}
\end{figure*}

To better understand the localization behavior of CROSS, we visualize cross-view matching and localization results on the KITTI and VIGOR datasets under unknown orientation, as shown in Fig.~\ref{fig:kitti_loc} and Fig.~\ref{fig:vigor_loc}.
On KITTI, Fig.~\ref{fig:kitti_loc}(a) and Fig.~\ref{fig:kitti_loc}(b) present successful cases with several prominent landmarks, where CROSS establishes reliable correspondences between the ground-view image and the satellite-view image.
More importantly, Fig.~\ref{fig:kitti_loc}(c) and Fig.~\ref{fig:kitti_loc}(d) show successful localization in more challenging open-road and natural scenes with fewer salient landmarks.
This indicates that the proposed 3D-grounded alignment and structure-aware matching can exploit structural relations among sparse cues, rather than relying solely on local appearance similarities.
The failed cases in Fig.~\ref{fig:kitti_loc}(e) and Fig.~\ref{fig:kitti_loc}(f) mainly occur in repetitive road scenes, where the exact position along the road-parallel direction is difficult to disambiguate from visual observations alone.
Nevertheless, the predicted orientation remains largely correct, suggesting that CROSS still captures the global directional structure of the scene.

On VIGOR, Fig.~\ref{fig:vigor_loc}(a) and Fig.~\ref{fig:vigor_loc}(b) show accurate localization in urban scenes with clear landmarks, such as buildings and intersections.
Fig.~\ref{fig:vigor_loc}(c) and Fig.~\ref{fig:vigor_loc}(d) further demonstrate that CROSS can localize in challenging scenes with fewer distinctive landmarks and repetitive patterns, such as trees, highlighting the importance of structural understanding under ambiguous visual evidence.
The failed cases in Fig.~\ref{fig:vigor_loc}(e) and Fig.~\ref{fig:vigor_loc}(f) occur in underground and indoor environments, where the ground-view observations have limited visible correspondences to the satellite map, making accurate cross-view localization nearly impossible.

\subsection{Semantic, Structural, and Geometric Learning}

In this subsection, we conduct both qualitative and quantitative analyses to investigate whether CROSS can learn stable semantics, reliable structure, and transferable geometry from cross-view localization.

\subsubsection{Stable Semantics}

\begin{figure*}
    \centering
    \includegraphics[width=0.95\linewidth]{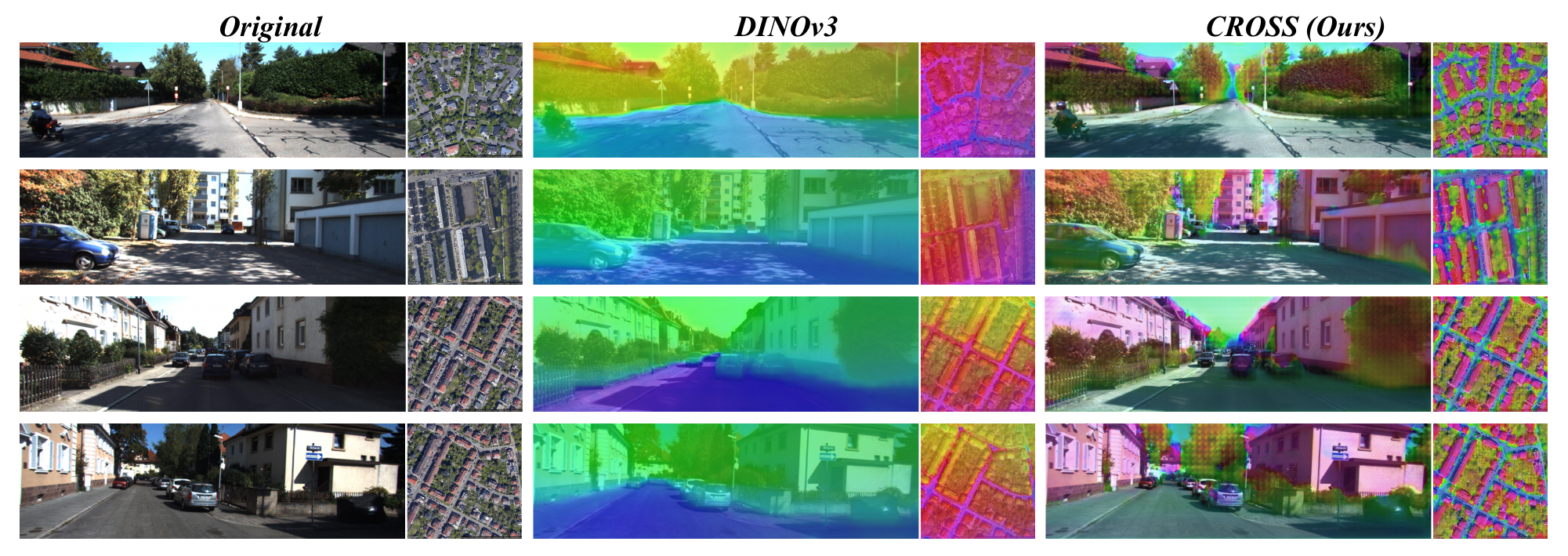}
    \caption{Visualization of semantic features predicted by DINOv3 and CROSS on the KITTI dataset. }
    \label{fig:kitti-semantic}
\end{figure*}

\begin{figure*}
    \centering
    \includegraphics[width=0.95\linewidth]{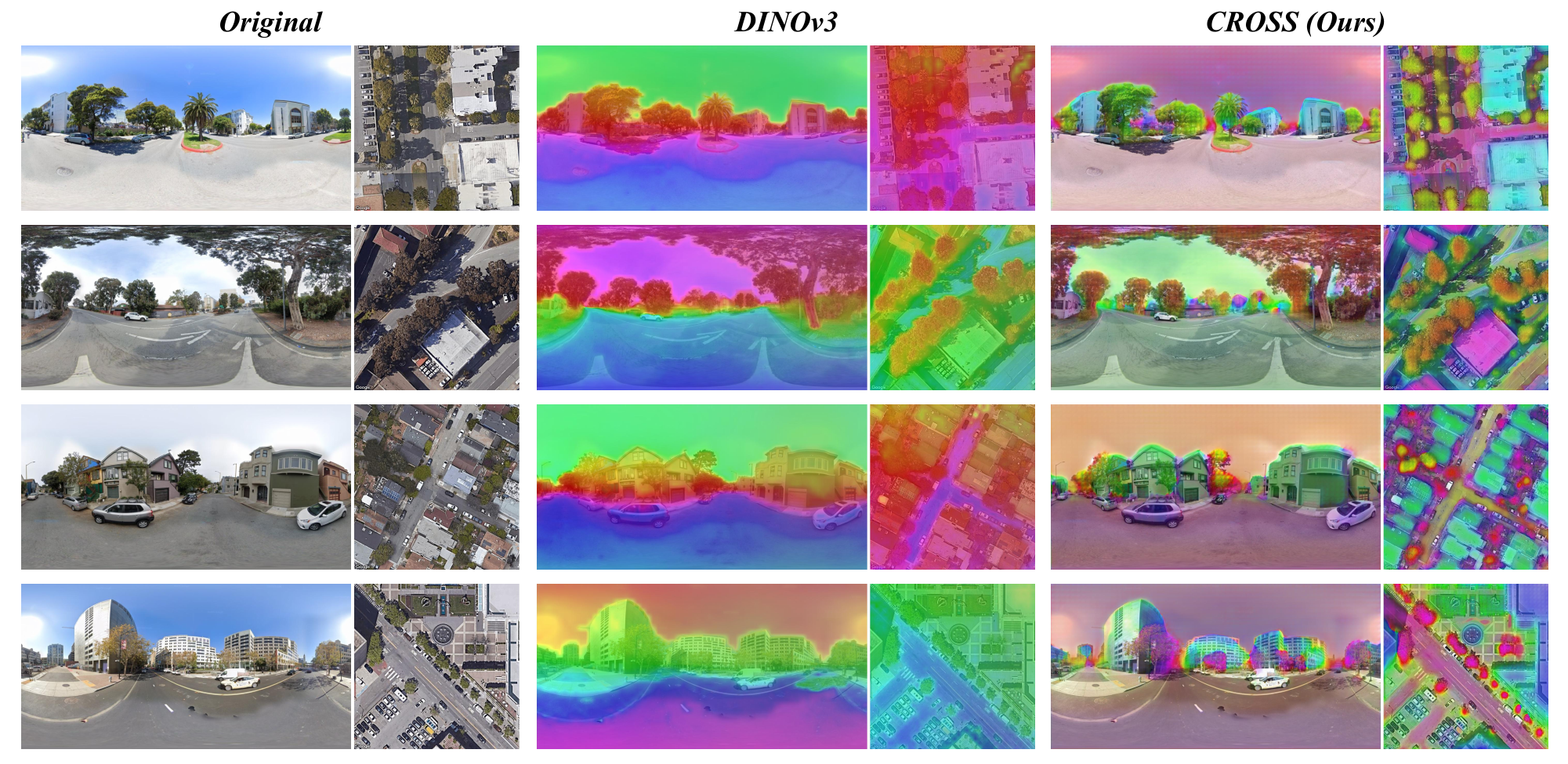}
    \caption{Visualization of semantic features predicted by DINOv3 and CROSS on the VIGOR dataset.}
    \label{fig:vigor-semantic}
\end{figure*}

To investigate the semantic representations learned by CROSS from cross-view localization, we compare the semantic features produced by CROSS and DINOv3.
Fig.~\ref{fig:kitti-semantic} and Fig.~\ref{fig:vigor-semantic} provide qualitative comparisons of cross-view semantic features on the KITTI and VIGOR datasets, respectively.
To ensure viewpoint-consistent visualization, we jointly fit PCA~\cite{pearson1901pca} on the ground-view and satellite-view feature maps and reduce the feature dimension to three RGB channels.
For CROSS, we further use the predicted reliability scores to adjust feature transparency, thereby highlighting reliable landmarks that contribute to localization.

The visualizations show that DINOv3 features exhibit limited cross-view consistency under large viewpoint gaps.
Semantically corresponding regions often produce inconsistent color responses between the ground and satellite views, making different areas difficult to align across viewpoints.
In contrast, CROSS learns more coherent and viewpoint-consistent semantic representations.
Major semantic regions, such as roads, buildings, and trees, are more clearly distinguishable within each view and exhibit more consistent color patterns across the two viewpoints.
This demonstrates that cross-view localization provides effective supervision for learning stable semantic representations under extreme viewpoint changes.

\begin{table}
    \caption{Cross-view semantic segmentation evaluation results on the VIGOR dataset. G denotes ground-view, S denotes satellite-view. The \protect\colorbox{bestcolor}{best performance} is highlighted.}
    \centering
    \begin{tabular}{ccccc}
    \toprule
    Methods & $\uparrow$mIoU & $\uparrow$Acc & $\uparrow$mIoU (G) & $\uparrow$mIoU (S) \\
    \midrule
    DINOv3 & 30.10 & 61.39 & 37.25 & 20.41\\
    CROSS & \best{54.55} & \best{69.40} & \best{48.58} & \best{60.61}\\
    \bottomrule
    \end{tabular}
    
    \label{tab:vigor-segmentation}
\end{table}

We also conduct a quantitative evaluation of unsupervised cross-view semantic segmentation on the VIGOR dataset.
We first use Mask2Former~\cite{cheng2022mask2former} and SparseMask~\cite{wu2019sparsemask,xia2023openearthmap} to generate segmentation masks for the two viewpoints, covering $C_s=4$ semantic classes: road, building, tree, and developed space.
Then we jointly apply PCA to the pixel-level features from both viewpoints, cluster the reduced features using $k$-means, and map each cluster assignment to its corresponding semantic class.
As shown in Tab.~\ref{tab:vigor-segmentation}, CROSS substantially improves semantic segmentation performance over DINOv3 in both viewpoints.
Specifically, CROSS increases the overall mIoU from 30.10 to 54.55 and the pixel accuracy from 61.39 to 69.40.
The improvement is observed for both ground-view and satellite-view imagery, with the corresponding mIoU increasing from 37.25 to 48.58 and from 20.41 to 60.61, respectively.
The particularly large improvement of 40.20 percentage points in satellite-view mIoU demonstrates the effectiveness of CROSS in the more challenging satellite-view setting, which contains denser object distributions, more diverse scene elements, and greater visual complexity than ground-view imagery.

\begin{table}
    \caption{Cross-view semantic consistency evaluation results on the VIGOR dataset. G denotes ground-view, S denotes satellite-view. The \protect\colorbox{bestcolor}{best performance} is highlighted.}
    \centering
    \begin{tabular}{ccccc}
    \toprule
         Methods & $\uparrow$R@1$_{G\rightarrow S}$ & $\uparrow$R@1$_{S\rightarrow G}$ & $\uparrow$ CSS \\
        \midrule
         DINOv3 & 54.74 & 56.03 & 0.3952 \\
         CROSS & \best{76.89} & \best{82.21} & \best{0.7211} \\
    \bottomrule
    \end{tabular}
    
    \label{tab:vigor-semantic}
\end{table}

We further evaluate whether these semantic representations remain consistent across viewpoints.
For each ground-satellite image pair, features belonging to the same semantic class are mean-pooled to obtain one prototype per class for each viewpoint.
We then compute a cosine similarity matrix $\mathbf{S}\in \mathbb{R}^{C_s \times C_s}$ between the ground-view and satellite-view prototypes.
Based on this similarity matrix, we report bidirectional Recall@1 (R@1) and the Cross-view Separation Score (CSS), which measures whether semantically corresponding prototypes are more similar across viewpoints than non-corresponding prototypes:

\begin{equation} 
\mathrm{CSS} = \frac{1}{C_s} \sum_{i=1}^{C_s} S_{ii} - \frac{1}{C_s(C_s-1)}\sum_{i=1}^{C_s}\sum_{\substack{j=1 \\ j\neq i}}^{C_s} S_{ij}. 
\end{equation}

As shown in Tab.~\ref{tab:vigor-semantic}, CROSS substantially outperforms DINOv3 across all cross-view semantic consistency metrics.
Specifically, CROSS achieves R@1 scores of 76.89 for ground-to-satellite retrieval and 82.21 for satellite-to-ground retrieval, improving over DINOv3 by 22.15 and 26.18 percentage points, respectively.
CROSS also increases the CSS score from 0.3952 to 0.7211, indicating that its semantic prototypes exhibit stronger cross-view correspondence while remaining more clearly separated from those of other semantic classes.
These results demonstrate that CROSS learns semantic representations that are substantially more discriminative and viewpoint-consistent than those produced by DINOv3.

\subsubsection{Reliable Structure}

\begin{figure*}
    \centering
    \includegraphics[width=0.95\linewidth]{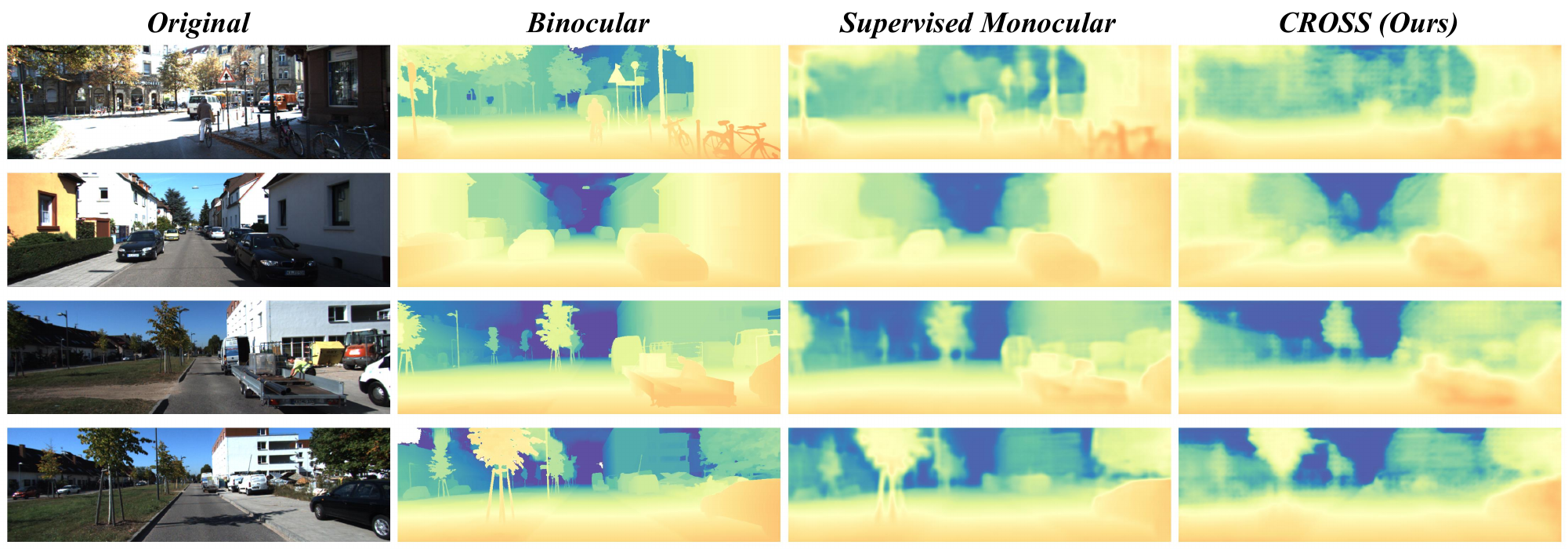}
    \caption{Comparison of structure learning on the KITTI dataset. The supervised monocular baseline adopts the same depth-prediction architecture as CROSS and is trained with an $\ell_1$ loss using binocular depth as supervision. CROSS recovers a similar metric structure to the supervised monocular baseline without direct supervision.}
    \label{fig:kitti-depth}
\end{figure*}

To further investigate whether CROSS learns reliable and scale-consistent structure from cross-view localization, we visualize its predicted depth maps in Fig.~\ref{fig:kitti-depth}.
Compared with the binocular depth reference and the directly supervised monocular baseline, CROSS recovers a similar global depth layout, including nearby road regions, intermediate objects such as vehicles and trees, and distant background structures.
Although the depth maps predicted by CROSS are smoother and less detailed around thin objects and object boundaries, they still preserve a coherent organization of scene structure and a consistent metric scale without relying on dense metric-scale supervision.
This behavior resembles perceptual-level 3D layout understanding, where fine-grained geometric details may be imperfect, but the overall spatial arrangement and scene-scale structure remain consistent.

\begin{table}
    \caption{Metric-scale depth map evaluation results on the KITTI dataset. The \protect\colorbox{bestcolor}{best performance} and the \protect\colorbox{secondcolor}{second-best performance} are highlighted.}
    \centering
    \begin{tabular}{cccc}
    \toprule
         Methods & $\downarrow$AbsRel & $\uparrow\delta1$ & $\uparrow\delta2$ \\
    \midrule
         DA3 & 0.285 & 0.286 & 0.801 \\
         CROSS (Weak) & \second{0.235} & \second{0.712} & \second{0.908} \\
         CROSS (Full) & \best{0.183} & \best{0.754} & \best{0.922} \\
    \bottomrule
    \end{tabular}
    
    \label{tab:kitti-depth}
\end{table}

Table~\ref{tab:kitti-depth} further quantitatively evaluates the metric-scale structure learned on the KITTI dataset, where binocular depth maps predicted by FoundationStereo~\cite{wen2025foundationstereo} are used as the reference.
Without direct metric-scale depth supervision, CROSS trained with 2-DoF weakly supervised cross-view localization substantially improves over Depth Anything 3 (DA3), reducing AbsRel from $0.285$ to $0.235$ and improving $\delta_1$ from $0.286$ to $0.712$.
With fully supervised cross-view localization, CROSS further improves all metrics, achieving the best AbsRel, $\delta_1$, and $\delta_2$.
These results demonstrate that CROSS can refine the scale-consistent structure understanding of current foundation models, such as Depth Anything 3, by learning from cross-view localization, even without direct metric-scale depth supervision.

\subsubsection{Transferable Geometry}

\begin{table*}[t]
\centering
\caption{Localization results across different viewpoints on the KITTI cross-area test set. The \protect\colorbox{bestcolor}{best performance} and the \protect\colorbox{secondcolor}{second-best performance} are highlighted.}
\begin{tabular}{lcccccccccc}
\toprule
& \multicolumn{5}{c}{Ground-View to Satellite-View} 
& \multicolumn{5}{c}{Ground-View to Ground-View} \\
\cmidrule(lr){2-6} \cmidrule(lr){7-11}
Method
& \multirow{2}{*}{Supervision}
& \multicolumn{2}{c}{$\downarrow$ Localization (m)} 
& \multicolumn{2}{c}{$\downarrow$ Orientation ($^\circ$)} 
& \multirow{2}{*}{Supervision}
& \multicolumn{2}{c}{$\downarrow$ Localization (m)} 
& \multicolumn{2}{c}{$\downarrow$ Orientation ($^\circ$)} \\
& 
& Mean & Median & Mean & Median 
& 
& Mean & Median & Mean & Median \\
\midrule

Loc$^2$
& \cmark
& \second{11.71} & \second{9.11} & 55.18 & \second{33.23} 
& \xmark
& 5.84 & 5.73 & 3.49 & 1.37 \\

DA3 
& \xmark
& 81.44 & 77.92 & \second{49.38} & 34.99 
& \cmark
& \best{0.67} & \best{0.44} & \best{0.19} & \best{0.08} \\

CROSS 
& \cmark
& \best{3.50} & \best{2.11} & \best{4.68} & \best{1.20}
& \xmark
& \second{1.01} & \second{0.63} & \second{0.62} & \second{0.31} \\

\bottomrule
\end{tabular}
\label{tab:kitti-geometry}
\end{table*}

\begin{figure*}[!h]
    \centering
    \includegraphics[width=0.95\textwidth]{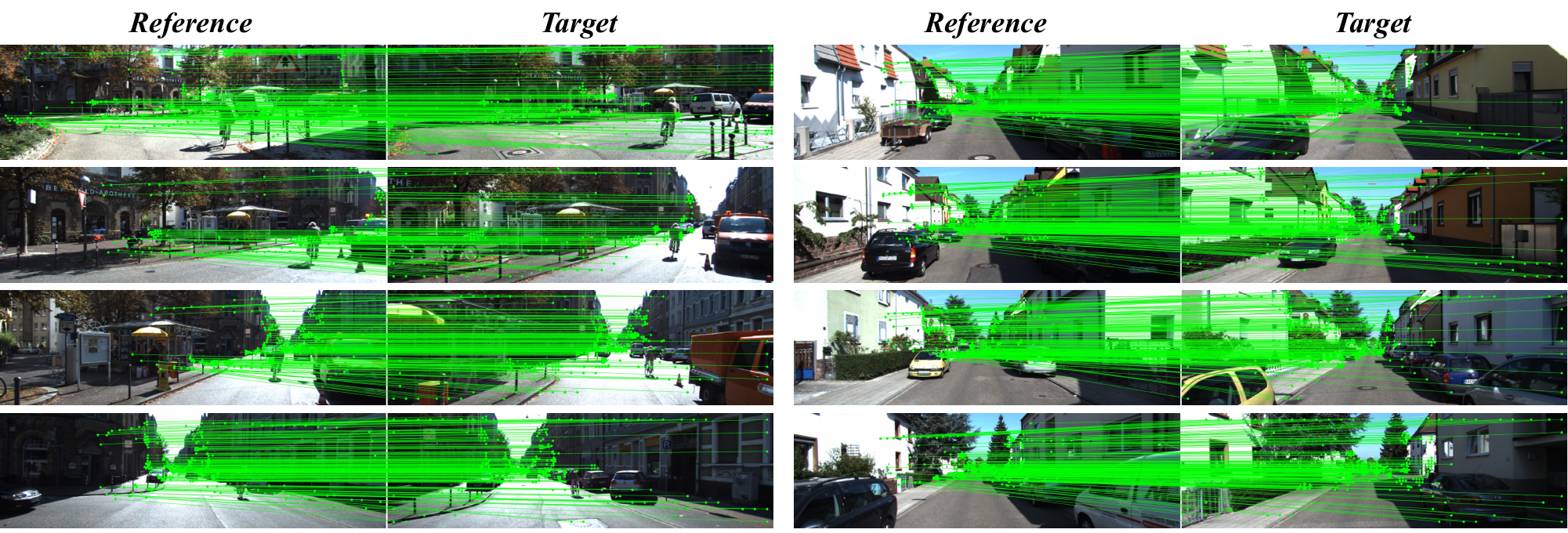}
    \caption{Visualization of ground-to-ground matching on two continuous trajectories from the cross-area test split of the KITTI dataset.}
    \label{fig:kitti-geometry}
\end{figure*}

To study whether CROSS learns transferable geometry from cross-view localization, we further evaluate the learned model on ground-to-ground matching, where the reference satellite map is replaced by a previous ground-view frame.
Since CROSS is trained only on the ground-to-satellite localization task, the key question is whether the geometry learned from this extreme cross-view transformation can transfer to another viewpoint relation without additional supervision.

As shown in Tab.~\ref{tab:kitti-geometry}, CROSS achieves the best performance on the ground-to-satellite localization task, where it is directly trained.
A notable observation is that Depth Anything 3 (DA3), despite showing impressive geometric understanding in ordinary scenarios with moderate viewpoint changes, fails under the extreme viewpoint gap of ground-to-satellite localization, resulting in a mean localization error of $81.44$ m.
In comparison, CROSS effectively learns geometry under large viewpoint changes, leading to robust performance in this challenging cross-view localization setting.

More importantly, CROSS also transfers well to ground-to-ground matching, despite receiving no supervision for this target transformation during training.
It achieves a mean localization error of $1.01$ m and a mean orientation error of $0.62^\circ$, substantially outperforming Loc$^2$, which is also trained for ground-to-satellite localization but fails to generalize well to this setting.
DA3 obtains the best ground-to-ground performance, benefiting from large-scale training on viewpoint transformations that are closer to this setting.
Nevertheless, CROSS achieves competitive results without such supervision, demonstrating that the geometry learned from cross-view consistency is not restricted to the training transformation.

The qualitative results in Fig.~\ref{fig:kitti-geometry} further show that CROSS produces coherent ground-to-ground correspondences.
For example, buildings and trees are correctly matched across different ground viewpoints under both straight and turning driving conditions.
This transferability is mainly enabled by the hypothesis ranking objective.
Instead of regressing an absolute pose target, CROSS evaluates candidate pose hypotheses according to their cross-view consistency.
As a result, the model is encouraged to learn general geometric reasoning from plausible alignments, rather than memorizing a specific transformation pattern, enabling transfer to unseen viewpoint relations.

\subsection{Ablation Study}

In this subsection, we conduct ablation studies to investigate the impact of the proposed modules on both cross-view localization and the learning of stable semantics, reliable structure, and transferable geometry.

\subsubsection{Ablation Study on Cross-View Localization}

\begin{table*}[t]
\centering
\caption{Ablation study results on the same-area split of the VIGOR dataset. The \protect\colorbox{bestcolor}{best performance} is highlighted.}
\begin{tabular}{ccccccc}
\toprule
3D-grounded &
Structure-aware &
\multirow{2}{*}{Hypothesis ranking} &
\multicolumn{2}{c}{$\downarrow$ Localization (m)} &
\multicolumn{2}{c}{$\downarrow$ Orientation ($^\circ$)} \\
\cmidrule(lr){4-5}
\cmidrule(lr){6-7}
alignment & matching & & Mean & Median & Mean & Median \\
\midrule
\xmark & \xmark & \xmark & 10.70 & 9.92 & 51.24 & 33.26\\
\cmark & \xmark & \xmark & 8.20 & 7.24 & 44.85 & 25.39\\
\cmark & \cmark & \xmark & 4.09 & 2.51 & 14.30 & 4.19\\
\cmark & \cmark & \cmark & \best{2.34} & \best{1.24} & \best{4.28} & \best{1.20}\\

\bottomrule
\end{tabular}
\label{tab:ablation}
\end{table*}

To evaluate how each module contributes to learning from cross-view localization, we conduct an ablation study on the same-area split of the VIGOR dataset, as shown in Tab.~\ref{tab:ablation}.
The baseline variant removes all proposed modules and performs direct 2D-2D matching followed by pose estimation, where the model is trained by regressing an absolute pose target.
This simple design yields a mean localization error of $10.70$ m and a mean orientation error of $51.24^\circ$, indicating that direct 2D-2D matching is insufficient under large viewpoint changes.

Introducing 3D-grounded alignment reduces the mean localization error from $10.70$ m to $8.20$ m and the mean orientation error from $51.24^\circ$ to $44.85^\circ$.
By constructing a 2D-3D-2D pathway, this module encourages the model to reason about reliable scene structure during cross-view localization.
The improvement suggests that 3D-grounded alignment provides more reliable candidate correspondences than direct 2D matching for subsequent pose estimation.
However, since this variant still mainly relies on point-to-point similarity, its performance remains limited when local appearances are ambiguous or semantically biased.

Adding structure-aware matching further improves the mean localization error to $4.09$ m and the mean orientation error to $14.30^\circ$.
Instead of enforcing isolated point-to-point correspondences, structure-aware matching aggregates matching evidence over the whole structure.
This design helps avoid representation collapse and instance-level bias, allowing semantic features to be learned from consistent structural context rather than individual local matches alone.
The substantial performance gain demonstrates that stable semantic learning benefits from evaluating cross-view consistency at the structure level.

Finally, the full CROSS model introduces hypothesis ranking, replacing direct pose regression with a ranking objective over candidate pose hypotheses.
This further reduces the mean localization error to $2.34$ m and the mean orientation error to $4.28^\circ$, achieving the best performance across all metrics.
Unlike regression-based learning from absolute pose targets, hypothesis ranking requires the model to compare the cross-view consistency of different pose hypotheses.
This encourages CROSS to learn transferable geometry that is not tied to a fixed transformation pattern, but instead emerges from reasoning over plausible cross-view alignments.

\subsubsection{Ablation Study on Semantic, Structural, and Geometric Learning}

We further conduct ablation studies to verify whether the proposed modules, including 3D-grounded alignment, structure-aware matching, and hypothesis ranking, enable the model to learn stable semantics, reliable structure, and transferable geometry.

\begin{table}
    \caption{Ablation study on cross-view semantic consistency on the VIGOR dataset. SAM denotes structure-aware matching. The \protect\colorbox{bestcolor}{best performance} is highlighted.}
    \centering
    \begin{tabular}{ccccc}
    \toprule
         Methods &  $\uparrow$R@1$_{G\rightarrow S}$ & $\uparrow$R@1$_{S\rightarrow G}$ & $\uparrow$ CSS \\
    \midrule
         CROSS & \best{76.89} & \best{82.21} & \best{0.7211} \\
         w/o SAM & 41.96 & 48.45 & 0.2027 \\
    \bottomrule
    \end{tabular}
    
    \label{tab:ablation-semantic}
\end{table}

For semantic learning, as shown in Tab.~\ref{tab:ablation-semantic}, removing structure-aware matching forces the model to rely on exclusive point-wise matching, where correspondences between semantically consistent but spatially different objects are still heavily penalized.
This leads to a substantial degradation in cross-view semantic consistency, with the cross-view semantic retrieval R@1 dropping from around 80 to below 50.
The CSS value also decreases, indicating that the learned semantic embeddings become less discriminative across different semantic classes.

For structural learning, we first remove the monocular-guided loss to examine its influence on structure prediction. 
As shown in Tab.~\ref{tab:ablation-structure}, removing only the monocular-guided loss mildly degrades the structure prediction accuracy, while the model can still infer relatively reliable structure through 3D-grounded alignment. 
We then remove the 3D-grounded alignment module, forcing the model to rely on 2D-2D matching followed by pose estimation. 
In this setting, the predicted structure is strongly constrained by the 2D matching results, since an incorrect correspondence may require an erroneous depth prediction to recover the correct camera pose. 
Without explicit 3D grounding, the structure head degrades significantly and fails to predict reliable scene structure, with the AbsRel error increasing from 0.183 to 0.792. 
This validates that 3D-grounded alignment is the key component for learning reliable structure, while the monocular-guided loss provides a beneficial structural prior.

\begin{table}[!t]
    \caption{Ablation study on metric-scale depth prediction on the KITTI dataset. Mono denotes monocular-guided loss. 3DGA denotes 3D-grounded alignment. The \protect\colorbox{bestcolor}{best performance} is highlighted.}
    \centering
    \begin{tabular}{cccc}
    \toprule
         Methods & $\downarrow$AbsRel & $\uparrow\delta1$ & $\uparrow\delta2$ \\
    \midrule
         CROSS & \best{0.183} & \best{0.754} & \best{0.922} \\
         w/o mono & 0.315 & 0.531 & 0.801\\
         w/o 3DGA & 0.792 & 0.001 & 0.003\\
    \bottomrule
    \end{tabular}
    
    \label{tab:ablation-structure}
\end{table}

\begin{table}[t]
    \centering
    \caption{Ablation study on zero-shot ground-to-ground localization on the KITTI dataset. The \protect\colorbox{bestcolor}{best performance} is highlighted.}
    \begin{tabular}{ccccc}
    \toprule
    \multirow{2}{*}{Methods} &
    \multicolumn{2}{c}{$\downarrow$ Localization (m)} &
    \multicolumn{2}{c}{$\downarrow$ Orientation ($^\circ$)} \\
    \cmidrule(lr){2-3}
    \cmidrule(lr){4-5}
    & Mean & Median & Mean & Median \\
    \midrule
    CROSS & \best{1.01} & \best{0.63} & \best{0.62} & \best{0.31}\\
    w/o HR & 4.09 & 3.89 & 27.39 & 18.70\\
    \bottomrule
    \end{tabular}
    \label{tab:ablation-geometry}
\end{table}

Finally, we replace the hypothesis ranking objective with an absolute regression objective to evaluate the learned geometry.
As shown in Tab.~\ref{tab:ablation-geometry}, when the model is trained with an absolute objective, it struggles to learn transferable geometry, since it may memorize the specific transformation observed during training rather than evaluating the cross-view consistency of different pose hypotheses.
Consequently, its zero-shot ground-to-ground localization performance drops sharply, with the localization error increasing from 1.01 m to 4.09 m and the orientation error increasing from 0.62$^\circ$ to 27.39$^\circ$.

Overall, these ablation studies validate the effectiveness of the three key modules in learning stable semantics, reliable structure, and transferable geometry from cross-view localization.

\subsection{Robustness Study}

\begin{table}[t]
\centering
\caption{Robustness study under different levels of initial translation noise on the same-area split of the VIGOR dataset.}
\begin{tabular}{ccccc}
\toprule
\multirow{2}{*}{Initial Noise (m)} &
\multicolumn{2}{c}{$\downarrow$ Localization (m)} &
\multicolumn{2}{c}{$\downarrow$ Orientation ($^\circ$)} \\
\cmidrule(lr){2-3}
\cmidrule(lr){4-5}
& Mean & Median & Mean & Median \\
\midrule
$\pm$10 & 1.91 & 1.13 & 3.77 & 1.18 \\
$\pm$20 & 2.34 & 1.24 & 4.28 & 1.20\\
$\pm$30 & 2.66 & 1.38 & 4.72 & 1.24 \\
$\pm$40 & 2.92 & 1.57 & 5.00 & 1.31 \\
\bottomrule
\end{tabular}
\label{tab:noise}
\end{table}

As CROSS supports weakly supervised training with noisy labels, we further investigate its robustness to the initial pose prior in this subsection.
In particular, we focus on initial translation noise, since the effect of different initial orientation noise levels has already been reported in Sec.~\ref{subsec:localization-results}.
As shown in Tab.~\ref{tab:noise}, CROSS remains stable as the initial translation noise increases from $\pm 10~\text{m}$ to $\pm 40~\text{m}$.
The mean localization error increases only mildly from $1.91~\text{m}$ to $2.92~\text{m}$, while the median error remains below $1.60~\text{m}$ even under the largest perturbation.
Meanwhile, the orientation error also changes only slightly, with the median error staying around $1.2^\circ$--$1.3^\circ$.
These results suggest that CROSS does not rely on an overly accurate initial pose prior.
Instead, it can recover reliable pose estimates from a relatively broad hypothesis region.

\subsection{Efficiency Study}
\label{subsec:efficiency}

\begin{table}[t]
\centering
\caption{Efficiency study on the same-area split of the VIGOR dataset. Res denotes the input resolution of the ground-view image, C2F denotes coarse-to-fine search, and FPS measures inference speed on an NVIDIA RTX 6000 Ada Generation GPU.}
\begin{tabular}{ccccccc}
\toprule
\multirow{2}{*}{Res} &
\multirow{2}{*}{C2F} & 
\multirow{2}{*}{$\uparrow$FPS} & 
\multicolumn{2}{c}{$\downarrow$ Loc (m)} &
\multicolumn{2}{c}{$\downarrow$ Ori ($^\circ$)} \\

\cmidrule(lr){4-5}
\cmidrule(lr){6-7}
& & & Mean & Median & Mean & Median \\
\midrule
\multicolumn{7}{l}{\textbf{\textit{CROSS (Ours)}}}\\
1022 & \xmark & 6.6 & 2.34 & 1.24 & 4.28 & 1.20\\
1022 & \cmark & 12.1 & 2.52 & 1.28 & 4.96 & 1.66\\
714 & \cmark & 14.2 & 2.56 & 1.29 & 5.19 & 1.70\\
504 & \cmark & 15.8 & 2.82 & 1.32 & 6.08 & 1.83 \\
\midrule
\multicolumn{7}{l}{\textbf{\textit{Loc$^2$}}}\\
1428 & - & 7.1 & 3.94 & 1.78 & 9.54 & 2.00\\
1022 & - & 8.3 & 7.32 & 4.16 & 31.50 & 7.43\\

\bottomrule
\end{tabular}
\label{tab:efficiency}
\end{table}

In this subsection, we investigate the efficiency of CROSS.
The proposed hypothesis ranking objective enables efficient training, since a small set of negative hypotheses is sufficient to guide the model to distinguish the ground-truth pose from perturbed alternatives.
During inference, evaluating more pose hypotheses with finer sampling can improve accuracy, but also increases computational cost.
To balance efficiency and effectiveness, we introduce a coarse-to-fine search strategy and compare CROSS with Loc$^2$ under different input resolutions.

Tab.~\ref{tab:efficiency} reports the efficiency results on the same-area split of the VIGOR dataset.
The coarse-to-fine search strategy significantly improves the inference efficiency of CROSS with only minor performance degradation.
At an input resolution of 1022, it increases the speed from 6.6 FPS to 12.1 FPS, while the mean localization error only changes from $2.34$ m to $2.52$ m and the mean orientation error from $4.28^\circ$ to $4.96^\circ$.
CROSS also shows strong robustness to resolution changes.
When the input resolution is reduced from 1022 to 714 and 504, the inference speed further increases to 14.2 FPS and 15.8 FPS, respectively, while the mean localization error only gradually increases from $2.52$ m to $2.56$ m and $2.82$ m, and the mean orientation error from $4.96^\circ$ to $5.19^\circ$ and $6.08^\circ$.
Compared with Loc$^2$, CROSS achieves a more favorable balance between accuracy and efficiency.
At the same resolution of 1022, CROSS runs faster than Loc$^2$, increasing the speed from 8.3 FPS to 12.1 FPS, while substantially reducing the mean localization error from $7.32$ m to $2.52$ m and the mean orientation error from $31.50^\circ$ to $4.96^\circ$.
Notably, even when CROSS uses a much lower resolution of 714 or 504, it still outperforms Loc$^2$ at a higher resolution of 1428 in both localization and orientation accuracy, while also achieving higher inference speed.
These results demonstrate that CROSS is not only more accurate and efficient than Loc$^2$, but also considerably more robust to input resolution changes.

\section{Conclusion}
In this work, we revisit cross-view localization as more than pose estimation and propose CROSS, a unified framework that learns stable semantics, reliable structure, and transferable geometry from cross-view localization.
Rather than formulating cross-view localization as 2D-2D matching followed by pose estimation, CROSS introduces 3D-grounded alignment to encourage reliable scene structure learning.
It further employs structure-aware matching to overcome the limitations of point-wise matching and learn semantic representations that remain stable across viewpoints.
The proposed hypothesis ranking objective supports both fully and weakly supervised training within a unified framework, while promoting geometric understanding that transfers to unseen viewpoint transformations.
Extensive experiments demonstrate that CROSS achieves state-of-the-art cross-view localization performance and, more importantly, effectively learns stable semantics, reliable structure, and transferable geometry.
In the future, we plan to further explore the potential of CROSS by scaling to larger and more diverse datasets, incorporating broader forms of extreme viewpoint variation, and extending cross-view localization into a general supervision signal for spatial representation learning.

\bibliographystyle{IEEEtran}
\bibliography{ref}

\end{document}